\documentclass[pmlr]{jmlr}% new name PMLR (Proceedings of Machine Learning)

\RequirePackage{graphicx}
 \usepackage{booktabs}
\usepackage{longtable}% for long tables
\makeatletter
\def\set@curr@file#1{\def\@curr@file{#1}} %temp workaround for 2019 latex release
\makeatother
\usepackage[load-configurations=version-1]{siunitx} % newer version

\theorembodyfont{\upshape}
\theoremheaderfont{\scshape}
\theorempostheader{:}
\theoremsep{\newline}

\jmlrvolume{252}
\jmlryear{2024}
\jmlrworkshop{Machine Learning for Healthcare}

\newcommand{\equal}[1]{{\hypersetup{linkcolor=black}\thanks{#1}}}

 \newcommand{\bs}[1]{\boldsymbol{#1}}        % boldsymbol

  \newcommand{\RR}[1]{\mathbb{R}^{#1}}
  \DeclareMathOperator*{\argmin}{argmin}
    \DeclareMathOperator*{\argmax}{argmax}
\title[Semi-Supervised Generative Models for Disease Trajectories]{Semi-Supervised Generative Models for Disease Trajectories: A Case Study on Systemic Sclerosis}
\author{%
\\
\Name{Cécile Trottet}\equal{These authors contributed equally.}
 \Email{cecileclaire.trottet@uzh.ch}\\ 
\addr
Department of Quantitative Biomedicine, University of Zurich, Zurich, Switzerland
\AND
\Name{Manuel Sch\"urch}\footnotemark[1] 
 \Email{manuel.schuerch@uzh.ch}\\
\addr
Department of Quantitative Biomedicine, University of Zurich, Zurich, Switzerland
\AND
\Name{Ahmed Allam}\\
\addr
Department of Quantitative Biomedicine, University of Zurich, Zurich, Switzerland
\AND
\Name{Imon Barua} \\
\addr  Department of Rheumatology, Oslo University Hospital,  University of Oslo, Oslo, Norway
\AND
\Name{Liubov Petelytska} \\
\addr Department of Rheumatology, University Hospital Zurich, University of Zurich, Zurich, Switzerland,  Department of Internal Medicine \#3, Bogomolets National Medical University, Kyiv, Ukraine
\AND
\Name{David Launay} \\
\addr Hôpital Huriez, CHU Lille, Lille University, Lille, France
\AND
\Name{Paolo Airò} \\
\addr Rheumatology and Clinical Immunology Unit, ASST Spedali Civili of Brescia, University of Brescia, Brescia, Italy
\AND
\Name{Radim Bečvář} \\
\addr Institute of Rheumatology, Department of Rheumatology, 1st Medical School, Charles University, Prague, Czech Republic
\AND
\Name{Christopher Denton} \\
\addr Centre for Rheumatology Royal Free, University College London Medical School, London, United Kingdom
\AND
\Name{Mislav Radic} \\
\addr Division of Rheumatology and Clinical Immunology, Department of Internal Medicine, University of Split, School of Medicine, University Hospital Center Split, Split, Croatia
\AND
\Name{Oliver Distler} \\
\addr Department of Rheumatology, University Hospital Zurich, University of Zurich, Zurich, Switzerland
\AND
\Name{Anna-Maria Hoffmann-Vold} \\
\addr Department of Rheumatology, University Hospital Zurich, University of Zurich, Zurich, Switzerland, Department of Rheumatology, Oslo University Hospital, Oslo, Norway 
\AND
\Name{Michael Krauthammer}\\
\addr Department of Quantitative Biomedicine, University of Zurich, Zurich, Switzerland
\AND
\Name{the EUSTAR collaborators}% \\
}

\begin{document}

\maketitle

\newpage
\begin{abstract}
  % Summary of the article.  Be sure to highlight how the work
  % contributes to our understanding of machine learning and healthcare.
We propose a deep generative 
approach using latent temporal processes for modeling and holistically analyzing complex 
disease trajectories, with a particular focus on Systemic Sclerosis (SSc). 
We aim to learn 
temporal latent representations of the underlying generative process that 
explain the observed patient disease trajectories 
in an interpretable and comprehensive way.
\\To enhance the interpretability of these latent temporal processes,
we develop a semi-supervised approach for 
disentangling
the latent space using established medical knowledge.
By combining the generative approach with medical definitions of different characteristics of SSc, we facilitate the discovery of
new aspects of the disease.
\\We show that the learned temporal latent processes can be utilized for further data analysis and clinical hypothesis testing, including finding similar patients and clustering SSc patient trajectories into novel sub-types.
Moreover, our method enables personalized online monitoring and prediction of multivariate time series with uncertainty quantification.
% We demonstrate the effectiveness of our approach in modeling systemic sclerosis,
% showcasing the potential of our generative model to capture 
% complex disease trajectories. 
\end{abstract}

\section{Introduction}

% XX is an important problem in machine learning and healthcare.  (Make
% sure that the clinicians can see the relevance! \emph{Unclear clinical
%   relevance is a major reason that otherwise strong-looking papers are
%   scored low/rejected.})

% Addressing this problem is challenging because XX.  (Make sure that
% you connect to the machine learning here.)  

% Others have tried, but XX remains tough.  (Acknowledge related work.)

% In this work, we...

% As you write, keep in mind that MLHC papers are meant to be read by
% computer scientists and clinicians.  In the later sections, you might
% have to use some medical terminology that a computer scientist may not
% be familiar with, and you might have to use some math that a clinician
% might not be familiar with.  That's okay, as long as you've done your
% best to make sure that the core ideas can be understood by an informed
% reader in either community.

\begin{figure}[htbp]

  \includegraphics[width=.7\linewidth]{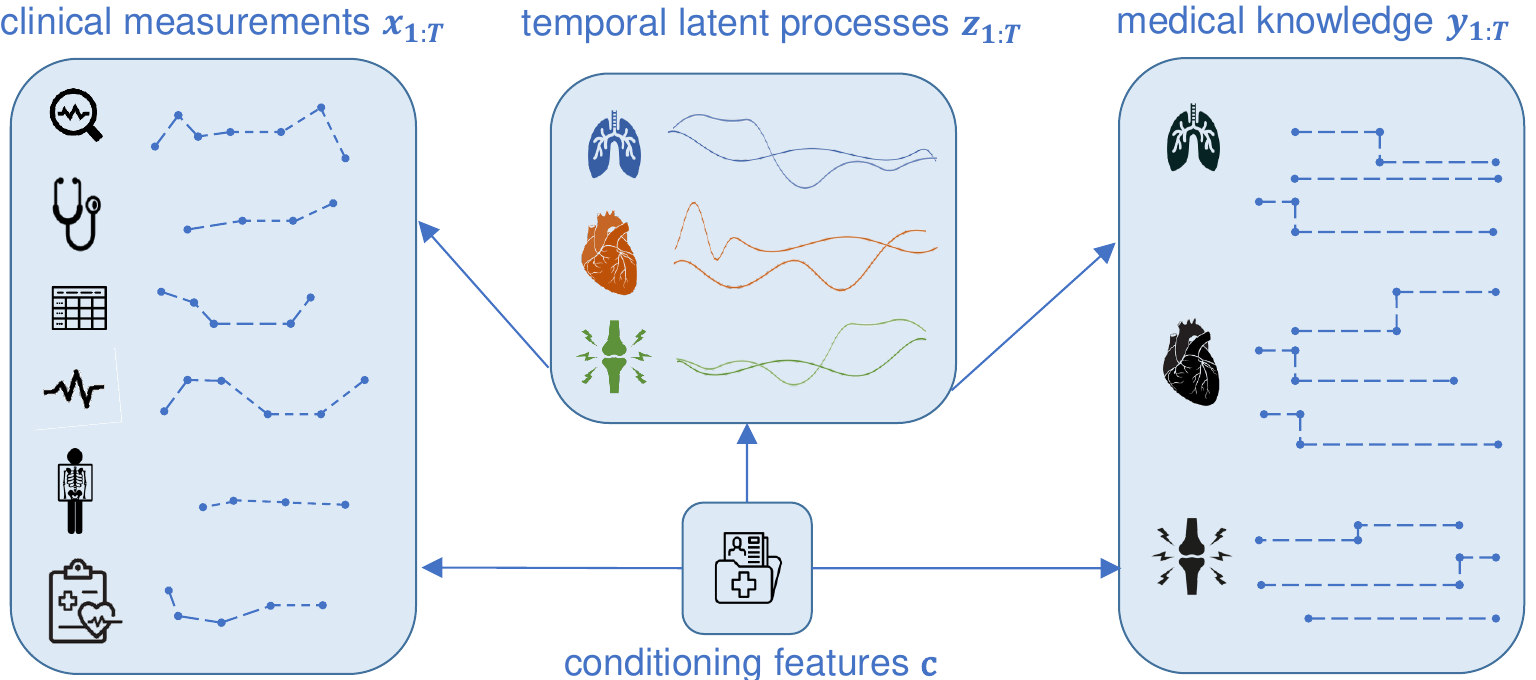}
  \centering
  \caption{Temporal generative model for systemic sclerosis. The latent temporal process $\bs{z}$ generates the observed $\bs{x}$ and $\bs{y}$ trajectories conditioned on data $\bs{c}$.}
  \label{fig:overview}
\end{figure}

\label{sec:introduction}
Understanding and analyzing 
clinical
trajectories of complex diseases, such as Systemic Sclerosis (SSc), is crucial for improving diagnosis, treatment, and patient outcomes \citep{allam2021analyzing}. However, modeling 
such
multivariate  time series data
poses significant challenges due to the high dimensionality of clinical measurements, low signal-to-noise ratio, 
sparsity, and the complex interplay of various potentially unobserved factors
influencing the disease progression \citep{allam2021analyzing}. 
Therefore, 
our primary goal 
is to develop 
a machine learning (ML) model suited for the holistic analysis of
temporal disease trajectories. 
Moreover, we aim to uncover meaningful temporal latent representations capturing the complex interactions within the raw data while also providing interpretable insights, 
and potentially revealing 
novel
medical aspects of clinical disease trajectories.
To achieve these goals, 
we present a deep generative temporal model that captures both the joint distribution of all observed longitudinal clinical variables and latent temporal variables (\autoref{fig:overview}).
 
Since inferring interpretable temporal representations in a fully unsupervised way is very challenging \citep{locatello2020sober}, 
we propose a semi-supervised approach for 
disentangling
the latent space using known medical knowledge
to enhance the interpretability. 
Combining an unsupervised latent generative model with known medical labels facilitates the discovery of novel medically-driven patterns in the data. 

Deep probabilistic generative models 
(\citeauthor{Tomczak2022DeepModeling} \citeyearpar{Tomczak2022DeepModeling})
provide a more holistic approach to modeling complex data than 
deterministic discriminative models.
By learning the joint distribution over all observed variables, they model the underlying data-generating mechanism.
In contrast, discriminative models only learn the conditional distribution of the target variable given the input variables. 

While our method is general and can be applied to a wide range of high-dimensional clinical datasets, in this paper, we demonstrate its effectiveness in
%focus on
modeling the progression of systemic sclerosis (SSc), a severe and yet only partially understood autoimmune disease. SSc triggers the immune system to attack the body's connective tissues, causing severe damage to the skin and multiple other internal organs. 
We seek to understand the evolution of SSc by modeling the patterns of organ involvement and progression. In doing so, we aim to learn temporal hidden representations that distinctly capture the disentangled medical disease processes related to each organ.

Our approach offers several 
contributions:
\begin{itemize}
\item \textbf{Interpretable Temporal Latent Processes:} 
Our generative model
allows the non-linear projection of  
patient trajectories onto lower-dimensional temporal latent processes, providing useful representations for the visualization and understanding of complex medical time series data.
\item \textbf{Semi-Supervised 
Guided 
Latent Processes:} 
To achieve more interpretable latent temporal spaces,
we propose a semi-supervised approach for 
disentangling 
the latent space with respect to medical knowledge.
  By combining the generative approach with medical domain knowledge, 
new aspects of the disease can be discovered.

\item
    \textbf{Online 
    Prediction with Uncertainty Quantification:}
    Our deep generative probabilistic model facilitates personalized online monitoring and reliable predictions of multivariate time series data with uncertainty quantification.
    
    \item \textbf{Facilitating Clinical Hypothesis Testing:} The learned temporal latent processes can be inspected for further data analysis and clinical hypothesis testing, such as finding similar patients and clustering the disease trajectories into new sub-types.

      \item \textbf{Large-Scale
      Analysis of Systemic Sclerosis
 %      \footnote{
 %      The focus of this paper is on the ML methodology, while  clinical  insights 
 % about
 % systemic sclerosis
 % will be discussed in a clinical follow-up paper, 
 % see \ref{se:clinic_ssc} for more details.
 %      }
      :}
      We demonstrate the potential of our ML model 
     for 
      comprehensively 
      analyzing 
      SSc for the first time at a large scale including multiple organs and various observed clinical variables.
\end{itemize}

\subsection*{Generalizable Insights about Machine Learning in the Context of Healthcare}

Our work offers contributions at the intersection of machine learning methodology and clinical practice, by proposing a new deep generative approach to model patient disease trajectories and conducting a large-scale ML analysis of organ involvement in SSc. We propose an approach to augment deep unsupervised generative models with medical knowledge, resulting in models with more interpretable and disentangled latent processes,  suited for further downstream tasks like clustering. Our work also highlights the challenges of developing a holistic end-to-end approach to process high-dimensional, sparse, and temporal clinical data.  Moreover, we demonstrate that our approach facilitates meaningful phenotyping of SSc and offers a novel methodology for understanding the disease's progression across multiple organs, setting a foundational baseline for future machine learning research on organ-specific disease modeling in SSc. While our expertise is suited to the modeling of SSc, we are confident that enriching deep generative models with targeted clinical knowledge holds potential for interdisciplinary collaborations between ML researchers and clinical experts to study various complex chronic diseases.  
% This section is \emph{required}, must keep the above title, and should
% be the final part of your introduction.  In about one paragraph, or
% 2-4 bullet points, explain what we should \emph{learn} from reading
% this paper that might be relevant to other machine learning in health
% endeavors.

% For example, a work that simply applies a bunch of existing algorithms
% to a new domain may be useful clinically but doesn't increase our
% understanding of the machine learning and healthcare; if that study
% also investigates \emph{why} different approaches have different
% performance, that might get us excited!  A more theoretical machine
% learning work may be in how it enables a new kind of clinical study.
% \emph{Reviewers and readers will look to evaluate (a) the significance
%   of your claimed insights and (b) evidence you provide later in the
%   work of you achieving that contribution}

\section{Related Work}

% Make sure you also put your work in the context of related
% work.  Who else has worked on this problem, and how did they approach
% it?  What makes your direction interesting or distinct?
\label{sec:background}

\subsection{Generative Latent Variable Models}
Learning latent representations from raw 
%high-dimensional
data has a long tradition in statistics and ML with foundational research such as principal component analysis \citep{hotelling1933analysis}, factor analysis \citep{lawley1962factor} or independent component analysis \citep{comon1994independent}, which all can be used to project high-dimensional
tabular data to a latent space.  
For temporal data, models with latent processes such as hidden Markov models \citep{baum1966statistical} %for discrete time 
%state-space-models or
and
Gaussian processes \citep{williams2006gaussian} %for continuous time 
have extensively been used for discrete and continuous time applications, respectively. 
%in many applications including 
%health-care data.
Conceptually, all these models can be viewed as probabilistic generative models with latent variables (e.g.\ \cite{murphy2022probabilistic}), however, these models only learn linear or simple relationships between the input data and the latent space.

In their seminal work on Variational Autoencoders (VAEs), \citet{kingma2013auto} proposed a powerful generalization for latent generative models.
The key idea is to use deep neural networks as function approximators to learn the moments of the data distribution, enabling the representation of arbitrarily complex distributions.
The parameters of the neural networks are inferred using amortized variational inference (VI) \citep{blei2017variational}, a powerful Bayesian inference method for approximating intractable probability distributions.
% Inference for the parameters of the neural networks is done with amortized variational inference (VI) \citep{blei2017variational}, a powerful approximate Bayesian inference tool.
There are various successors building and improving on the original model, for instance, conditional VAE \citep{sohn2015learning}, LVAE
\citep{sonderby2016ladder}, or VQ-VAE \citep{van2017neural}. Moreover, 
there are also several extensions that explicitly model time in the latent space such as RNN-VAE \citep{chung2015recurrent}, 
GP-VAE \citep{casale2018gaussian, fortuin2020gp}, or longitudinal VAE \citep{ramchandran2021longitudinal}.

While these approaches have showcased remarkable efficacy in generating diverse objects such as images or modeling time series, the interpretability of the resulting latent spaces or processes remains limited for complex data. Moreover, the true underlying distributions of known processes often cannot be recovered, and instead become \textit{entangled} within a single latent factor \citep{bengio2013representation}. 
%For instance,  
Thus, there is ongoing research in designing generative models with disentangled latent factors, such as
 $\beta-$VAE \citep{higgins2016beta}, factorVAE \citep{kim2018disentangling}, TCVAE \citep{chen2018isolating} or temporal versions including
 disentangled sequential VAE \citep{hsu2017unsupervised} and
 disentangled GP-VAE \citep{bing2021disentanglement}.

However, learning interpretable and disentangled latent representations is highly difficult or even impossible for complex data without any inductive bias 
\citep{locatello2020sober}. Hence, purely unsupervised modeling falls short, leading researchers to focus on weakly supervised latent representation learning instead
\citep{locatello2020weakly, zhu2022sw, palumbo2023deep}.
In a similar spirit, we tackle the \emph{temporal} semi-supervised guidance of the latent space by using sparse labels representing established medical domain knowledge. We model the progression of complex diseases
in an unsupervised way using the raw temporal clinical measurements, while augmenting the model with temporal medical labels. 

\subsection{Analyzing Disease Trajectories with ML}
Recently, extensive research has focused on modeling and analyzing clinical time series with machine learning --
%for which 
we refer to \citet{allam2021analyzing} for an overview. However, most approaches focus on deterministic time series forecasting, and only a few focus on interpretable representation learning with deep models \citep{Trottet2023ExplainableDiseases} and irregularly sampled times \citep{chen2023dynamic} or on online uncertainty quantification with generative models 
\citep{schurch2020recursive, cheng2020sparse, rosnati2021mgp}.

A few approaches aim at uncovering disease stages from electronic health records in a fully unsupervised way \citep{yang2014finding, wang2014unsupervised, alaa2019attentive} or with a self-supervised approach \citep{raghu2023sequential}. However, as motivated in the previous section, and by \cite{chen2021probabilistic}, we rather develop a semi-supervised approach to model latent disease stages 
%with a generative approach 
using sparse medical labels. 

Recent approaches for clustering time series
\citep{lee2020temporal, srivastava2023expertnet, qin2023t} focus on learning predictive embeddings for future events. However, these techniques are not designed for semi-supervised environments with high-dimensional, multi-labeled, and sparse temporal data. While \cite{noroozizadeh2023temporal} and \cite{holland2023clustering} leverage contrastive learning to cluster time series,  we rather adopt a generative approach to fully model the complete patient trajectory.
%raghu2023sequential: self-supervised
%lee2020temporal: predictive clustering
%qin2023t: predictive clustering

Furthermore, prior research on data-driven analysis of systemic sclerosis is limited.
In their recent review, \citet{bonomi2022use} discuss the existing studies applying machine learning for precision medicine in systemic sclerosis. However, all of the listed studies are limited by the small cohort size (maximum of 250 patients), making the use of deep learning models challenging. Deep models were only used for analyzing imaging data, mainly related to nailfold capillaroscopy \citep{garaiman2022vision}. Furthermore, most existing works solely focus on the involvement of a single organ in SSc, namely interstitial lung disease (ILD), and on forecasting methods \citep{bonomi2022use}.  To the best of our knowledge, our work is the first attempt at a comprehensive  and large-scale (N=5673 patients)
ML analysis 
of
systemic sclerosis involving multiple organs and a wide range of observed clinical variables together with a systematic integration of the latest medical knowledge.

\section{Methods}

% Tell us your techniques!  If your paper is develops a novel machine
% learning method or extension, then be sure to give the technical
% details---as you would for a machine learning publication---here and,
% as needed, in appendices.  If your paper is developing new methods
% and/or theory, this section might be several pages.

% If you are combining existing methods, feel free to cite other
% packages and papers and tell us how you put them together; that said,
% the work should stand alone for someone in that general machine
% learning area.  

% \emph{Lack of technical details, such that the soundness of the
%   methods can be verified, is a major reason that otherwise
%   strong-looking papers are scored low/rejected.}
\label{sec:method}
We analyze patient medical histories that consist of two main types of data: raw temporal clinical measurements $\bs{x} = \bs{x}_{1:T}\in \RR{D \times T}$, such as 
%heart rate,
blood pressure,
and sparse medical knowledge labels $\bs{y} = \bs{y}_{1:T}\in \RR{P \times T}$, describing the medical definitions of selected aspects of the disease, such as the medical definition of severity staging of the heart involvement in SSc (\autoref{fig:overview}).
%We denote by $\tau$ the time points at which $x$ and $y$ are recorded. 
The medical knowledge definitions (Appendix \ref{sec:app:def}) are typically derived from multiple clinical measurements using logical operations. For example, a patient may be classified as having ``lung involvement" if certain conditions are satisfied, for instance, $\bs{x}^{(i)}>\varepsilon$ OR $\bs{x}^{(j)} = 1$. 
Both the raw measurements and labels are irregularly sampled, and we denote by $\bs{\tau}_{1:T} \in \RR{T}$ the vector of observation time-points of $\bs{x}$ and $\bs{y}$.
Moreover, there is non-temporal information denoted by $\bs{s}\in \RR{S}$ such as patient demographics, alongside additional temporal covariates such as medications $\bs{p}_{1:T} \in \RR{P \times T}$  for each patient.

We condition our generative model on the context variable $\bs{c} = \{\bs{\tau}, \bs{p}, \bs{s} \}$ 
%to 
%be able to generate latent processes under certain conditions, for instance when a specific medication is administered. 
to take into account the heterogeneous patient preconditions.
Furthermore, in the next sections, we introduce our approach to learning unobserved multivariate latent processes denoted as $\bs{z} = \bs{z}_{1:T}\in \RR{L \times T}$, responsible for generating both the raw clinical measurements $\bs{x}_{1:T}$ and the medical labels $\bs{y}_{1:T}$. Specifically, we use the different temporal medical labels to disentangle the $L$ dimensions of the latent processes by allocating distinct dimensions to represent different medical knowledge labels.

We assume a dataset  $\{\bs{x}^i_{1:T_i}, \bs{y}^i_{1:T_i}, \bs{c}^i_{1:T_i} \}_{i=1}^N$ of $N$ patients, 
and omit the dependency to $i$  and the time index when the context is clear. 
Note that the measurements and medical labels are often partially observed, see more details in 
%Section 
Appendix \ref{sec:partially_obs}. A table of the main introduced symbols is provided in \autoref{tab:symbols} in the appendix.

\subsection{Generative Model}
\label{subsec:gen_model}
\begin{figure*}[htbp]
    \floatconts
    {fig:model_combined}
    {\caption{Semi-supervised temporal latent variable model. The left panel shows the model architecture with the inference and generative components, and the right panel describes the guidance networks. We have independent guidance networks for each medical label, taking as input a subset of the latent dimensions and predicting the corresponding medical label.}}
    {%
         \raisebox{5em}{\subfigure[Model architecture.]{\label{fig:model}%
            \includegraphics[width=0.6\linewidth]{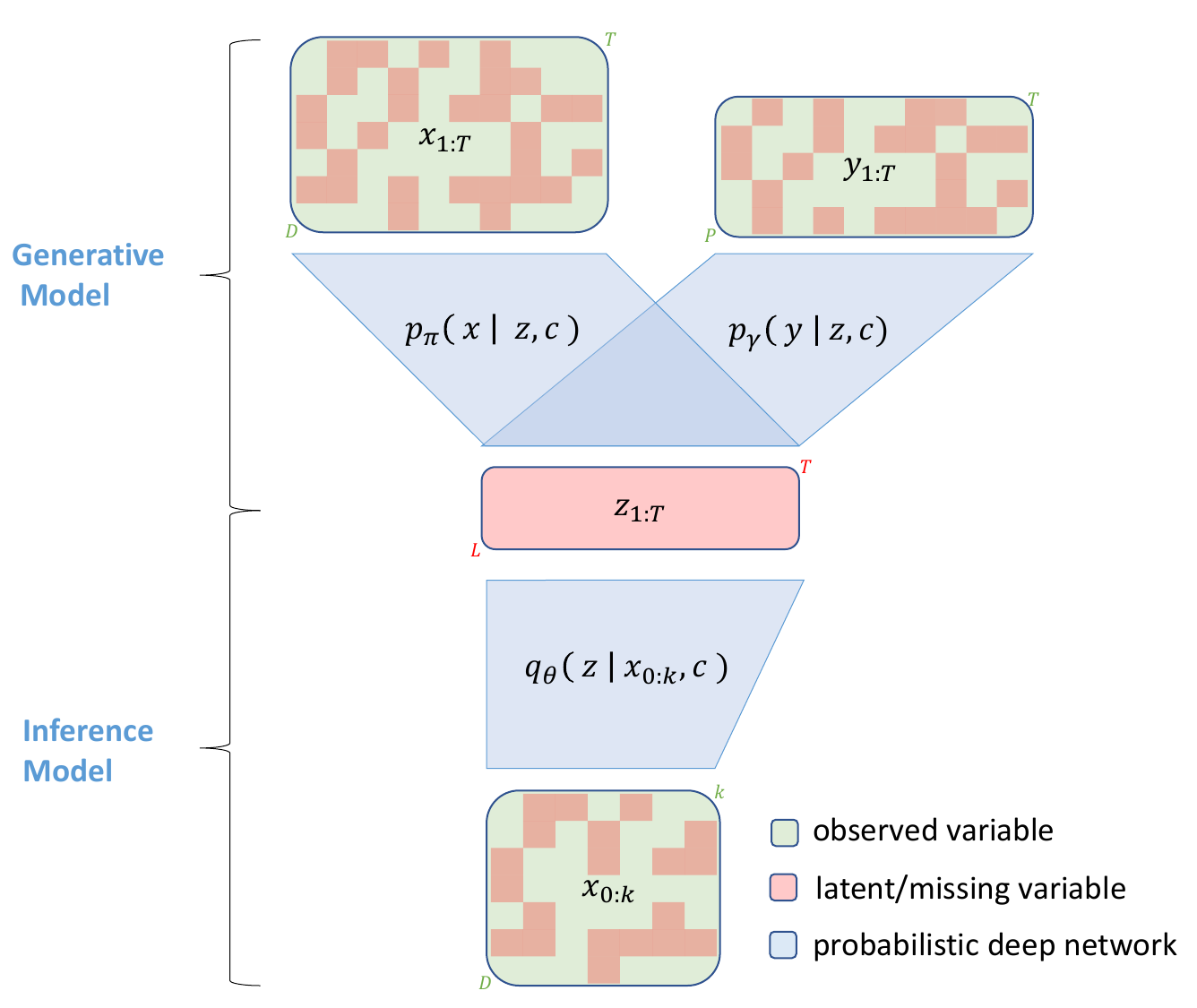}}}%
        \qquad% You may adjust the horizontal space if necessary
        
        \subfigure[Guidance networks.]{\label{fig:model_guid}%
            \includegraphics[width=0.9\linewidth]{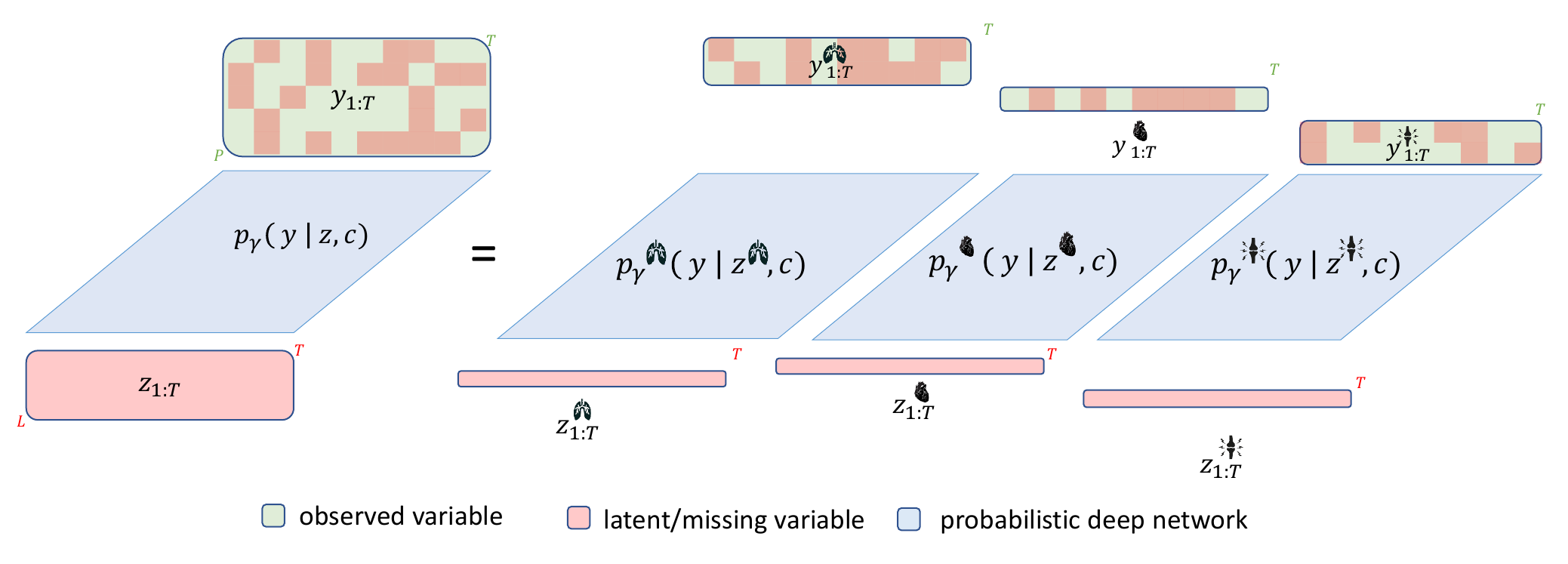}}}
    \end{figure*}

%For modeling multi-variate time series, 
We propose the probabilistic conditional generative latent variable model
\begin{align*}
    p_{\psi}(\bs{x}, \bs{y}, \bs{z} \vert \bs{c})
    =
    p_{\pi}(\bs{x} \vert \bs{z}, \bs{c})
     p_{\gamma}(\bs{y} \vert \bs{z}, \bs{c})
      p_{\phi}(\bs{z} \vert \bs{c}),
\end{align*}
with 
learnable
%latent 
prior network 
$ p_{\phi}(\bs{z} \vert \bs{c})$, 
%concept likelihood 
measurement likelihood network
$p_{\pi}(\bs{x} \vert \bs{z}, \bs{c})$,
and 
%disentangling
guidance
networks 
$ p_{\gamma}(\bs{y} \vert \bs{z}, \bs{c})$,
 where 
$\psi = \{\gamma, \pi, \phi\}$ are learnable parameters (\ref{fig:model}). We assume conditional independence of $\bs{x}$ and $\bs{y}$ given $\bs{z}$ and $\bs{c}$.
%
%of the neural networks. 
%
Although the measurements and the medical labels are conditionally independent, 
%given the latent variables, 
the marginal distribution 
$p_{\psi}(\bs{y}, \bs{x}\vert \bs{c}) = \int p_{\psi}(\bs{y}, \bs{x}, \bs{z} \vert \bs{c}) d \bs{z}$ allows arbitrarily rich correlations among the observed variables.
For the sake of brevity, we do not include the time index explicitly.
\subsection{Prior of Latent Process}
\label{sec:priorL}
%\subsubsection{Agnostic Model}
We use a learnable prior network  for the latent temporal variables $\bs{z} \in \RR{L \times T}$,
% depending on the  conditional factors 
%  $\bs{c}=\{\bs{\tau}, \bs{p}, \bs{s}\}$, 
 that is,
\begin{align*}
 p_{\phi}(\bs{z} \vert \bs{c})
&
  =
 \prod_{t=1}^T 
     \prod_{l=1}^L
   \mathcal{N}\left(
   \bs{z}_t^l \vert 
\mu_{\phi}^l(\bs{c}_t), \sigma^l_{\phi}(\bs{c}_t)
   \right),
 \end{align*}
 conditioned on the context variables 
 $\bs{c}=\{\bs{\tau}, \bs{p}, \bs{s}\}$,
 so that  time-varying or demographic effects can be learned in the prior
 (Appendix 
\ref{sec:app:prior}).
 %in order to represent the heterogeneous trajectories.
%
The means $\mu_{\phi}^l(\bs{c}_t)$ and variances $\sigma^l_{\phi}(\bs{c}_t)$ are parametrized by deep neural networks.
We assume a factorized Gaussian prior distribution per time and latent dimensions, 
 however, many interesting extensions 
% to this simple model 
 including  continuous-time priors
 %for instance GP-VAE
  %\citep{CasaleGaussianAutoencoders, FortuinGP-VAE:Imputation} 
 are straightforward 
 (Appendix \ref{sec:diff_prior}).% for more details.
% \begin{figure}[htbp]
% \floatconts
%   {fig:model}
%   {\caption{Semi-supervised temporal latent variable model with generative and inference model.
%  % \rs{version 3,4, or 5?}
%   }}
%     {\includegraphics[width=.9\linewidth]{graphs/model.png}}
% \end{figure}
\subsection{Likelihood of Measurements}
\label{sec:likelihood}
The probabilistic likelihood network 
%(aka decoder) 
maps the 
%$L$ 
latent temporal processes $\bs{z} \in \RR{L \times T}$ together with the context variables $\bs{c}$ to the
%concepts $\bs{y}$ and 
clinical
measurements $\bs{x} \in \RR{D \times T}$, i.e.
we assume the following factorization
\begin{align*}
 p_{\pi}(\bs{x} \vert \bs{z}, \bs{c})
 &
 =
 \prod_{t=1}^T 
 %\left\{
  \prod_{d \in \mathcal{G}}
\mathcal{N}(x_{t}^d \vert \mu_{\pi}^d
%(\bs{z}_{t}, \bs{c}_{t})
,  \sigma^d_{\pi}
%(\bs{z}_{t}, \bs{c}_{t})
)
\prod_{d \in \mathcal{K}}
\mathcal{C}(x_{t}^d \vert p_{\pi}^d
%(\bs{z}_{t}, \bs{c}_{t})
),
% \right\}
\end{align*}
where we assume time and feature-wise conditional independence.
We assume either Gaussian $\mathcal{N}$ or categorical $\mathcal{C}$ likelihoods for the observed variables $\bs{x}$, where $\mathcal{G}$ and $\mathcal{K}$ are the corresponding indices. 
The moments of these distributions are parametrized by deep neural networks, i.e.\ the
   mean $\mu_{\pi}^d=\mu_{\pi}^d(\bs{z}_{t}, \bs{c}_{t})$, variance    $\sigma^d_{\pi}=\sigma^d_{\pi}(\bs{z}_{t}, \bs{c}_{t})$, and
   category
  probability vector 
  $p_{\pi}^d=p_{\pi}^d(\bs{z}_{t}, \bs{c}_{t})$.
  %, respectively.%,  obtained from the deep neural network.
  Although the likelihood is a parametric distribution, the posterior distribution can be arbitrarily complex after marginalizing out the latent process $\bs{z}$.

\subsection{Semi-Supervised Guidance Network}
\label{sec:meth:dis}
We propose a semi-supervised approach to disentangle the latent process $\bs{z}$ with respect to defined medical labels $\bs{y} = \bs{y}_{1:T}\in \RR{P \times T}$. 
In particular, we assume
\begin{align*}
p_{\gamma}(\bs{y} \vert \bs{z}, \bs{c})
& 
 =
 \prod_{t=1}^T 
  \prod_{g=1}^G
   \mathcal{C}(y_{t}^{\nu(g)} \vert h_{\gamma}^{\nu(g)}( \bs{z}_{t}^{\varepsilon(g)}, \bs{c}_t ) ),
\end{align*}
where $\vert G \vert$ is the number of different medical labels. We assume categorical distributions for all medical labels, but the extension to continuous labels is straightforward. $h_{\gamma}^{\nu(g)}(\bs{z}_{t}^{\varepsilon(g)}, \bs{c}_t )$ is a deep parametrized category probability matrix, 
and $\nu(g)$ and $\varepsilon(g)$  correspond to the indices of the  $g$th  guided medical label, and the  indices in the latent space defined for guided label $g$,
%organ $o$, 
respectively (Figure \ref{fig:model_guid}). 

% \begin{figure*}[htbp]
% \floatconts
%   {fig:model_guid}
%   {\caption{Guidance networks. We define separate independent networks for each medical label.
%   }}
%     {\includegraphics[width=0.7\linewidth]{graphs/pdfresizer.com-pdf-crop (2).pdf}}
% \end{figure*}

\subsection{Posterior of Latent Process}
\label{sec:post}
We are mainly interested in the posterior distribution $p_{\psi}(\bs{z} \vert \bs{x}, \bs{y}, \bs{c})$ of the latent process given the observations,
which we 
%will 
approximate with
%We introduce 
an 
amortized 
variational distribution %$q_{\theta}(\bs{z} \vert \bs{x}, \bs{c})$,
(Section \ref{sec:inference}, Appendix \ref{sec:app:inference})
$$ q_{\theta}(\bs{z} \vert \bs{x}, \bs{c})  \approx  p_{\psi}(\bs{z} \vert \bs{x}, \bs{y}, \bs{c}).$$
%so that 
%where we assume 
%where 
We use the 
%following 
amortized variational distribution 
%(sometimes called encoder)
\begin{align*}
%q_{\theta}(\bs{z} \vert \bs{x}, \bs{c})
%=
q_{\theta}(\bs{z} \vert \bs{x}_{0:k}, \bs{c})
&
=
\prod_{t=1}^T
\prod_{l=1}^L  \mathcal{N}( z_{t}^l\vert \mu_{\theta}^l(\bs{x}_{0:k}, \bs{c}), \sigma^{l}_{\theta}(\bs{x}_{0:k}, \bs{c}) ) 
 \end{align*}
 with variational parameters $\theta$ and $0 \leq k \leq T$.
 %, where $\bs{x}_{1:0} = \varnothing$. 
 Note that only the measurements $\bs{x}_{0:k}$ until observation $k$ are part of the variational distribution, and not the medical labels $\bs{y}$. If $k=T$, there is no forecasting, whereas for $0\leq k< T$, we can also forecast the future latent variables $\bs{z}_{k+1:T}$ from the first measurements $\bs{x}_{0:k}$. 

\subsection{Probabilistic Inference} 
\label{sec:inference}
Since exact inference with the 
marginal likelihood
 $ p_{\psi}(\bs{x}, \bs{y} \vert \bs{c})
 = \int
  p_{\gamma}(\bs{y} \vert \bs{z}, \bs{c})
       p_{\pi}(\bs{x} \vert \bs{z}, \bs{c})
      p_{\phi}(\bs{z} \vert \bs{c})
      d\bs{z}$
 is not feasible (Appendix \ref{sec:app:inference}),
 we apply amortized variational inference \citep{blei2017variational} by maximizing a
lower bound 
$\log p_{\psi}(\bs{x}, \bs{y}\vert \bs{c}) \geq \mathcal{L}( \psi, \theta; \bs{x},  \bs{y}, \bs{c} )$
of the intractable marginal log likelihood. %$ \log p_{\psi}(\bs{x}, \bs{y} \vert \bs{c})$,
For a fixed $k$, this leads to the following objective function
%that is,   
\begin{align}
  \label{eq:ELBO}
    \begin{split}
 \mathcal{L}_k
  ( \psi, \theta; \bs{x},  \bs{y}, \bs{c})
  =
 & ~
 \mathbb{E}_{q_{\theta}(\bs{z} \vert \bs{x}_{0:k}, \bs{c})}\left[
 \log p_{\pi}(\bs{x} \vert \bs{z}, \bs{c}) 
 %p_{\gamma}(\bs{y} \vert \bs{z}, \bs{c})
  \right]
  \\
  +
  &~
\alpha ~\mathbb{ E}_{q_{\theta}(\bs{z} \vert \bs{x}_{0:k}, \bs{c})}\left[
 \log 
 %p_{\pi}(\bs{x} \vert \bs{z}, \bs{c}) 
 p_{\gamma}(\bs{y} \vert \bs{z}, \bs{c})
  \right]
  \\
  -
  &~
 \beta ~KL\left[
q_{\theta}(\bs{z} \vert \bs{x}_{0:k}, \bs{c} )
~\vert\vert~
p_{\phi}(\bs{z} \vert \bs{c})
  \right]
  ,
   \end{split}
\end{align}
where we introduce weights $\alpha$ and  $\beta$ inspired by the disentangled $\beta-$VAE \citep{higgins2016beta}.
The first term $\mathbb{E}_{q_{\theta}(\bs{z} \vert \bs{x}_{0:k}, \bs{c})}\left[
 \log p_{\pi}(\bs{x} \vert \bs{z}, \bs{c}) 
  \right]$
  is unsupervised, whereas the second
  $$\alpha \mathbb{E}_{q_{\theta}(\bs{z} \vert \bs{x}_{0:k}, \bs{c})}\left[
 \log 
 %p_{\pi}(\bs{x} \vert \bs{z}, \bs{c}) 
 p_{\gamma}(\bs{y} \vert \bs{z}, \bs{c})
  \right]$$
  is supervised 
  and 
  $\beta KL\left[
q_{\theta}(\bs{z} \vert \bs{x}_{0:k}, \bs{c} )
\vert\vert
p_{\phi}(\bs{z} \vert \bs{c})
  \right]$
is a regularization term, ensuring that 
the posterior is close 
to the prior with respect to the
Kullback-Leibler (KL) divergence.
Since all dimensions in the latent space $\bs{z}$ are connected to all the measurements $\bs{x}$ through the likelihood network,
all the potential correlations between clinical measurement variables can be exploited in an unsupervised fashion while disentangling the latent variables using the guidance networks for $\bs{y}$. 
The expectation over the variational distribution $\mathbb{E}_{q_{\theta}(\bs{z} \vert \bs{x}_{0:k}, \bs{c})}$ is approximated with a few Monte-Carlo samples (Appendix \ref{sec:app:inference}).
%and KL is the Kullback-Leibler divergence.

Given a dataset with $N$ $\mathrm{iid}$ patients $\{\bs{x}_{1:T_i}^i, \bs{y}_{1:T_i}^i, \bs{c}_{1:T_i}^i\}_{i=1}^N$,  the optimal 
%generative and variational 
parameters are obtained by the  maximization task
$$\psi^*, \theta^* = 
\argmax_{\psi, \theta}
\sum_{i=1}^N
\sum_{k=0}^{T_i}
\mathcal{L}_k( \psi, \theta; \bs{x}^i,  \bs{y}^i, \bs{c}^i ),
$$
which is solved with stochastic optimization using mini-batches of patients and different values for $k$ (Appendix \ref{sec:app:Nsamples}).
%(see more details in ).
Since real-world time series data often contains many missing values, the objective function can be adapted accordingly (Appendix 
\ref{sec:partially_obs}).
\subsection{Online Prediction with Uncertainty Quantification}
\label{sec:monit}
Our model can be used for online monitoring and continuous prediction of high-dimensional medical label and clinical measurement distributions  
based on an increasing number of available past clinical observations $\bs{x}_{0:k}$ for $k=0,1,\ldots, T$. The distributions
\begin{align*}
q_*
%_{\gamma^*, \theta^*}
(\bs{y} \vert \bs{x}_{0:k}, \bs{c} ) 
&=
\int
p_{\gamma^*}(\bs{y} \vert \bs{z}, \bs{c} ) q_{\theta^*}(\bs{z} \vert \bs{x}_{0:k}, \bs{c} ) d \bs{z}
\\
q_*
%_{\pi^*, \theta^*}
(\bs{x} \vert \bs{x}_{0:k}, \bs{c} ) 
&=
\int
p_{\pi^*}(\bs{x} \vert \bs{z}, \bs{c} ) q_{\theta^*}(\bs{z} \vert \bs{x}_{0:k}, \bs{c} ) d \bs{z}
\end{align*}
are approximated 
with two-stage Monte-Carlo sampling (Appendix \ref{sec:app:predictive_distribution}).
 The former can be used to automatically label and forecast the multiple medical labels based on the raw and partially observed measurements, whereas the latter corresponds to the reconstruction and forecasting of partially observed clinical measurement trajectories.
 Note that these distributions represent a complex class of potentially multi-modal distributions.

\subsection{Patient Similarity and Clustering}
\label{subsec:patient_similrity}
The learned posterior network
$q_{\theta^*}(\bs{z}_{1:T} \vert \bs{x}_{1:T}, \bs{c}_{1:T} )$ can be used to map any
%temporal 
observed
patient 
trajectory 
$\mathcal{T}_i = \{\bs{x}_{1:T_i}^i, \bs{c}_{1:T_i}^i\}$ 
to their latent trajectory
$$\mathcal{H}_i
=
h(\mathcal{T}_i)
=
\mathbb{E}_{
q_{\theta}(\bs{z}_{1:T_i}^i \vert \bs{x}_{1:T_i}^i, \bs{c}_{1:T_i}^i)
}
\left[
 \bs{z}_{1:T_i}^i
  \right]
$$
by taking the mean of the latent process. 
These temporal latent trajectories $\{ \mathcal{H}_i \}_{i=1}^N$ of the $N$ patients in the cohort are used to define a patient similarity over the partially observed and high-dimensional original disease trajectories
$\{ \mathcal{T}_i \}_{i=1}^N$.
Through our semi-supervised generative approach, the latent trajectories effectively capture the important components from $\bs{x}_{1:T_i}^i$ and $\bs{y}_{1:T_i}^i$, without explicitly depending on $\bs{y}_{1:T_i}^i$. Indeed, all the information related to the medical labels is learned by $\theta$. 

Since defining a patient similarity measure between two trajectories $\mathcal{T}_i$ and $\mathcal{T}_j$ in the original space is very challenging, due to the missingness and high dimensionality of the variables, we instead define it in the latent space, setting 
$$
d_{\mathcal{T}}\left(
\mathcal{T}_i, \mathcal{T}_j
\right)
=
d_{\mathcal{H}}\left(
\mathcal{H}_i, \mathcal{H}_j
\right).
$$
To measure the similarity $d_{\mathcal{H}}\left(
\mathcal{H}_i, \mathcal{H}_j
\right)$ between latent trajectories, we employ the \emph{dynamic-time-warping (dtw)} measure to account for the different lengths of the trajectories as well as the potentially misaligned disease progressions in time \citep{muller2007dynamic}. 
We then utilize the similarity measure to cluster the disease trajectories and identify similar patient trajectories
as discussed in Section
\ref{sec:clustering}.
\subsection{Deep Probabilistic Networks}
\label{sec:modelchoice}

As shown in Figures \ref{fig:model} and \ref{fig:model_guid}, our model combines several deep probabilistic networks. For the posterior $q_{\theta}(\bs{z} \vert \bs{x}_{0:k}, \bs{c})$, we implemented a temporal network with fully connected and LSTM layers \citep{hochreiter1997long}  and multilayer perceptrons (MLPs) for the prior $p_{\phi}(\bs{z} \vert \bs{c})$, guidance $p_{\gamma}(\bs{y} \vert \bs{z},\bs{c})$ and likelihood $p_{\pi} (\bs{x} \vert \bs{z}, \bs{c})$ networks. Implementation details are provided in Appendix \ref{sec:app:archi}.

By omitting the guidance  $p_{\gamma}(\bs{y} \vert \bs{z}, \bs{c})$ or likelihood networks  $p_{\pi}(\bs{x} \vert \bs{z}, \bs{c})$, we recover well-established temporal latent variable models. Specifically, removing the guidance networks transforms the model into a deterministic predictive LSTM-Autoencoder, or probabilistic predictive LSTM-VAE if we learn the latent space distribution. Moreover, if we exclude the likelihood network $p_{\pi} (\bs{x} \vert \bs{z}, \bs{c})$, the model operates in a fully supervised setting, focusing solely on optimizing the latent space for the prediction of the medical labels $\bs{y}$. 
%Furthermore, the likelihood variance can either be learned, or kept constant as is common practice \citep{rybkin2021simple}. 
% We evaluated the predictive performance of the guided model in the probabilistic and deterministic settings, with or without learning the likelihood variance.
Many further architectural choices could be explored, such as a temporal likelihood network or a Gaussian process prior (Appendix \ref{sec:diff_prior}), but they are beyond the scope of this paper. 

\section{Cohort}
We evaluate our model on the European Scleroderma Trials and Research (EUSTAR) database. The EUSTAR database extensively documents organ involvement in SSc for about 20’000 patients. For a detailed description of the database, we refer the reader to \citet{meier2012update, hoffmann2021progressive}. We use this database because this work is part of a broader initiative aiming to find the optimal medical definitions for organ involvement in SSc, leveraging data from the EUSTAR registry. 

We included $5673$ patients with at least $5$ and at most $15$ medical visits. We used $6$ static variables related to the patients' demographics and almost $40$ clinical measurement variables, mainly related to the lung, heart, and joint monitoring in SSc (Appendix \ref{app:model_variables}).
\subsection{Data extraction and Feature Choices}
The clinical measurement variables and patient demographics are directly available in the EUSTAR database. We provide a list of the used clinical measurement variables in Appendix \ref{app:model_variables}. They were selected based upon clinical relevance for modeling SSc. Each medical label is based on multiple EUSTAR variables (cf definitions in \ref{sec:app:def}) and created by using logical operations. For instance, the lung is involved if $\text{ILD on HRCT}\footnote{Interstitial Lung Disease on High-Resolution computed Tomography} = \text{YES}$ OR $\text{FVC}\footnote{Forced Vital Capacity} < 70 \%$. 
\subsection{Missing values}
Missingness is a common issue in medical records. We used mean value imputation for missing clinical measurements. However, we did not train our model to reconstruct these missing measurements, i.e.\ the imputed values are not part of the optimized loss (cf Appendix \ref{sec:partially_obs}), thus mitigating the bias induced by the missingness. Additionally, given that the medical labels often rely on multiple EUSTAR variables, they are even sparser due to a propagation of the missingness. The advantage of our semi-supervised approach is that it relies solely on available labels for guidance, without necessitating any label imputation.
\\

\noindent
The code and examples using an artificial dataset are available as supplementary material. 

\section{Results on the EUSTAR Database} 

% \emph{Depending on the claim you make in the paper, different
%   components may be important for this section.}

\subsection{Study Design: Modeling Systemic Sclerosis} 
% Before jumping into the results: what exactly are you evaluating?
% Tell us (or remind us) about your study design and evaluation
% criteria.
\label{sec:SSc}
We aim to model the overall SSc disease trajectories as well as the distinct organ involvement trajectories for patients from the EUSTAR database. 
%We provide a description of the database in Appendix \ref{sec:app:data}. 

We focus on the involvement of three important organs in SSc, namely the lung, heart, and arthritis in the joints.  Each organ has two related medical knowledge labels: \emph{involvement} and \emph{stage}. Based upon the medical definitions provided in Appendix \ref{sec:app:def}, for each of the three organs, we created labels signaling the organ involvement (yes/no) and severity stage ($1-4$), respectively.  We write $o(m)$, $m\in \{involvement, stage\}:=\mathcal{M}$, $o \in \mathcal{O} : = \{lung, heart, joints\}$, to refer to the corresponding medical label for organ $o$. We project the $D=34$ and $P=11$ input features to a latent process $\bs{z}$ of dimension $L=21$. 

For each organ, we guide a distinct subset of $7$ latent processes (non-overlapping subsets), thus all of the dimensions in $\bs{z}$ are guided (\ref{fig:model_guid}).
Following the notations from Section \ref{sec:meth:dis}, we assume the following guidance network structure
\begin{align*}
p_{\gamma}(\bs{y} \vert \bs{z}, \bs{c})
& =
 %p_{\gamma}(\bs{y}_{1:T} \vert \bs{z}_{1:T}, \bs{c}_{1:T})
% =
 \prod_{t=1}^T 
  \prod_{\substack{o \in \mathcal{O}}}
  \prod_{\substack{m \in \mathcal{M}}}%\\{\{inv}, \\{stage\}}\\}}
 p_{\gamma}(\bs{y}_{t}^{\nu(o(m))} \vert \bs{z}_{t}^{\varepsilon(o(m))}, \bs{c}_t),
  %\\
\end{align*}
where $\nu(o(m))$ and $\varepsilon(o(m))$ are the corresponding indices of the dimensions in the output and latent process, respectively. 
\subsubsection{Evaluation}
Given the conditioning data $\bs{c}$ and the clinical measurements $\bs{x}_{0:k}$ up to a given time-step $k$, our model learns the optimal parameters of the variational distribution of the complete latent trajectory $q_{\theta}(\bs{z} \vert \bs{x}_{0:k}, \bs{c})$, of the likelihood $p_{\pi} (\bs{x} \vert \bs{z}, \bs{c})$ and of the guidance networks $p_{\gamma}(\bs{y} \vert \bs{z},\bs{c})$. Thus given $\bs{x}_{0:k}$ and $\bs{c}$, our model predicts the complete trajectories of both $\bs{x}$ and $\bs{y}$. 

We aim to propose a holistic model that prioritizes versatility over achieving cutting-edge predictive performance. Thus, we evaluate the predictive trade-offs of our approach versus well-established deterministic and probabilistic temporal deep latent variable models optimized separately for each of the predictive tasks, i.e.\ predicting only $\bs{x}$ or $\bs{y}$ in a fully supervised way.

We evaluate the interpretability and disentanglement of the latent processes in our model against fully unsupervised methods. Furthermore, we evaluate and discuss the clinical relevance of the trajectory clusters identified for SSc patients.     
Lastly, we follow an index patient to showcase how our model enhances the understanding of their disease course, including online patient monitoring, patient trajectory sampling, and visualizations in the latent space. We refer the reader to the Appendix \ref{sec:app:res} for additional results related to patient similarity, adjustment of uncertainty quantification to out-of-distribution data, and sampling of prior trajectories.  

\subsection{Predictive Performance Evaluation} 

% Present your numbers and be sure to compare proposed methods against appropriate baselines. 
% You should provide a summary of
% the results in the text, 
% as well as in tables (such as
% Table~\ref{tab:example}) and figures (such as
% Figure~\ref{fig:example}).  
% You may use subfigures/wrapfigures 
% so that figures don't have to span the whole page or multiple figures are side by side.

% \begin{table}[t]
%   \centering 
%   \caption{Description with the main take-away point. Note that the caption should appear \emph{above} the table.}
%   \begin{tabular}{llll}
%   \toprule
%     \textbf{Method} & \textbf{Metric1} & \textbf{Metric2} & \textbf{Metric3} \\
%     \midrule
%     Baseline & 1.1 & 2.3 & 0.1 \\ 
%     NetNet & 41.3 & 31.9 & 77.4 \\ 
%     \bottomrule
%   \end{tabular}
%   \label{tab:example} 
% \end{table}

%  \begin{figure}[t]
%    \centering 
%    \includegraphics[width=2.5in]{plot.pdf} 
%    \caption{Description with the main take-away point. Note that figure captions should appear below the figure.}
%    \label{fig:example} 
%  \end{figure} 
\subsubsection{Baselines}
\label{seq:res:bsl}
The temporal baselines follow a similar encoder-decoder architecture as our model and are optimized to predict either $\bs{x}$ or $\bs{y}$ as targets. Similarly to our model, their temporal encoders take as input $\bs{x}_{0:k}$ and $\bs{c}$ and learn the distribution of the latent variables $\bs{z}$. The predictive decoders take as input a sampled $\bs{z}$ and predict the future targets. We implemented probabilistic and deterministic versions of each baseline. The encoders of the probabilistic models learn the mean and variance of the distributions of the latent variables, and the encoders of the deterministic models only learn their mean.

Similarly to our model, the temporal encoders contain LSTM and fully connected layers, and the decoders are MLPs. We denote LSTM-MLP-x and LSTM-MLP-y for the deterministic supervised models trained to predict $\bs{x}$ or $\bs{y}$, respectively, and LSTM-MLP-x* and LSTM-MLP-y* for the probabilistic variants. 
% The supervised baseline models are obtained by excluding either the likelihood or guidance networks and thus optimize simplified loss functions.
% Like our model, they take as input $\bs{x}_{0:k}$ and $\bs{c}$ and have the same latent space size. 
We expect these models to generally outperform our approach, since they have a similar model capacity but learn simpler tasks. Their training objective can be expressed as simplified versions of our model's objective (Equation \eqref{eq:ELBO}). The associated loss functions can be found in Appendix \ref{app:res:bsl}.
%We found that models without any bottleneck layers achieved similar performances, and thus for the sake of simplicity, we only report here the results for the bottleneck architectures. 

In addition to these temporal deep learning models, we also evaluated our approach against a non-temporal MLP baseline taking as input the conditioning data and the last available values of each clinical measurement $\bs{x}$ before the prediction time-point. We denote this baseline as MLP-xy, as it predicts both $\bs{x}$ and $\bs{y}$. Lastly, we also implemented a naive cohort baseline drawing a value from the empirical distribution of the variable in the cohort (assuming a Gaussian distribution for continuous variables and a categorical distribution otherwise).
% Furthermore, we implemented two non-ML-driven baselines, one individualized and one cohort-based. The individualized baseline predicts the patient's last available measurement for a variable as its future value. The cohort baseline predicts a value sampled from the empirical distribution of the variable in the cohort (assuming a Gaussian distribution for continuous variables and a categorical distribution otherwise). 
We used $5-$fold cross-validation 
to select the hyperparameters that achieved the lowest validation loss for each model. Details about the inference process are provided in \autoref{sec:inference} and Appendix  \ref{sec:app:optim}. 

\subsubsection{Results}
% For the prediction of the medical labels $\bs{y}$, we compared our model to two fully supervised models, i.e.\ where the latent space $\bs{z}$ is only optimized for the prediction of the medical labels. These  LSTM-based models are obtained by discarding the likelihood network, and thus unlike our model do not learn the distribution of $\bs{x}$. We trained a probabilistic variant, learning the distribution of the latent space $\bs{z}$, and a deterministic variant, learning a deterministic $\bs{z}$.
In \autoref{fig:performance}, we report the predictive performance of the different models for the prediction of future $\bs{x}$ and $\bs{y}$ versus time to prediction. We report the average macro F1 score for categorical variables and the mean absolute error (MAE) for continuous variables. 

In the first panel of \autoref{fig:performance}, we evaluate the models' performance for the prediction of medical labels $\bs{y}$. Both of the task-specific models, i.e.\ the LSTM-MLP-y (yellow) and LSTM-MLP-y* (orange), slightly outperform our model (red), as expected since they only have to learn one category of outcomes. Furthermore, our model outperforms the MLP-xy (grey) and naive (green) baselines. We report the performance results separately for each medical label in Appendix \ref{sec:app:res}. 

% For the prediction of the clinical measurements $\bs{x}$, we compared our model's performance to two models obtained by discarding the guidance networks, resulting in LSTM-(variational) autoencoder architectures. The LSTM-VAE learns the distributions of $\bs{z}$ and $\bs{x}$ and the LSTM-AE is optimized only to predict the distribution of $\bs{x}$. Thus, unlike our model, both of these baselines do not learn the distribution of $\bs{y}$.
The last two panels of \autoref{fig:performance} show the prediction performance for categorical and continuous $\bs{x}$ versus time to prediction. For categorical $\bs{x}$, our model (red) performs similarly or outperforms all of the models, except for the first time-step, where the MLP-xy baseline (grey) performs the best. For continuous features, the LSTM-MLP-x (purple) outperforms our model in terms of MAE. There is no significant difference between our model and the LSTM-MLP-x* (brown), even though our model also learns the distribution of $\bs{y}$. Lastly, our model greatly outperforms the MLP-xy (grey) and naive (green) baselines.

As expected, the task-specific deterministic models generally slightly outperform our model, when allowed a similar capacity in the latent space, since they have to learn simpler tasks and fewer variables. In \autoref{fig:perf_latent} in Appendix \ref{sec:app:res}, we show that by decreasing the capacity of the LSTM-MLP-x, via reduction of the latent space dimension, we recover a similar performance to our multi-task holistic model.

% As expected, most of the task-specific supervised models slightly outperform our multi-task holistic approach. 

% As described in \autoref{sec:modelchoice}, we assessed the model's predictive performance in probabilistic and deterministic settings, and with either learning the likelihood network variance $\sigma^*$ or setting $\sigma = 1$.
\begin{figure}[htbp]
\floatconts
{fig:performance}
{\caption{Model performances versus time to prediction for $\bs{x}$ and $\bs{y}$ forecasting. 
}}
{\includegraphics[width=0.65\linewidth]{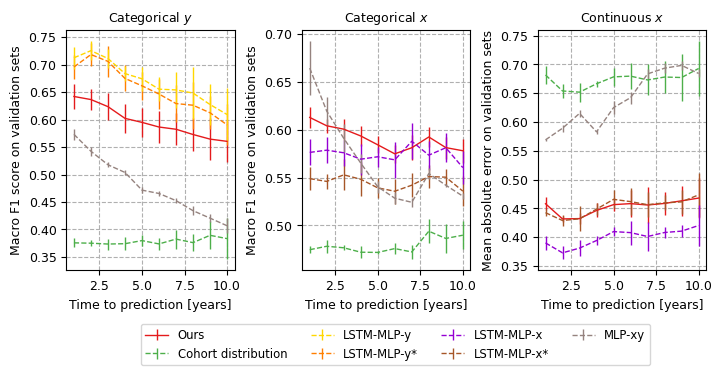}}
\end{figure}

% \autoref{fig:performance} shows the performance of predicting the clinical measurements $\bs{x}$. All of the deep learning models greatly outperform non-ML-driven individualized or cohort-based baselines. 
% The individualized baseline predicts the patient's last available measurement for a variable as its future value. The cohort baseline predicts a value sampled from the empirical Gaussian/Categorical distribution of the variable in the cohort. The models with learned $\sigma^*$ perform slightly better at predicting continuous $\bs{x}$ while enforcing $\sigma = 1$ allows the models to better learn the categorical $\bs{x}$. The same holds for the prediction of the categorical medical labels $\bs{y}$ (\autoref{fig:perf_y} in the appendix). Furthermore, there is no significant decrease in performance in probabilistic versus deterministic settings, even though an additional regularization term is optimized (\autoref{sec:modelchoice}). 
% In Appendix \ref{sec:app:res}, we additionally compare the performance for $\bs{x}$ prediction of our guided model versus the optimal unguided baseline. 

To evaluate the uncertainty quantification, we computed the coverage of the forecasted $95\%$ confidence intervals (CI) for continuous variables and the calibration for categorical variables. Furthermore, we computed the average ratio between CI length and feature range versus time to prediction. CIs are on average wider for long-term predictions and out-of-distribution data points (\autoref{fig:ood} in Appendix \ref{sec:app:res}). 
% For continuous $\bs{x}$ forecasting, both probabilistic models achieve coverage of $92 \pm 1\%$ and of $98 \pm 0\%$ for the deterministic models, thus all slightly diverging from the optimal $95 \%$. 
For continuous $\bs{x}$ forecasting, our model and the LSTM-MLP-x* achieve coverage of $92 \pm 1\%$ both, and the LSTM-MLP-x of $98 \pm 0\%$, thus all slightly diverging from the optimal $95 \%$.
All of the models have accurate calibration for categorical $\bs{x}$ and $\bs{y}$ forecasting, as shown in \autoref{fig:calib} in Appendix \ref{sec:app:res}.

% The probabilistic model with learned $\sigma^*$ strikes the best balance between predictive capabilities, coverage and generative ability. 

% In the next sections, we explore further applications and results of our model. While the performance was computed on validation sets, the subsequent results are derived from applying our model to a separate withheld test set. Furthermore, all of the $t-$SNE projections \citep{van2008visualizing} of the test set were obtained following the procedure described in Appendix \ref{sec:app:tsne}.
\subsubsection{Online Prediction with Uncertainty Quantification}
\begin{figure}[htbp]
    \floatconts
    {fig:monit_combined}
    {\caption{Online monitoring for $p_{\text{idx}}$. The model uses information prior to the dashed line as input and predicts the values after.}}
    {%
        \subfigure[Predicted probabilities of organ involvement. The heatmap reflects the predicted probabilities.]{\label{fig:inv_pat}%
            \includegraphics[width=0.35\linewidth]{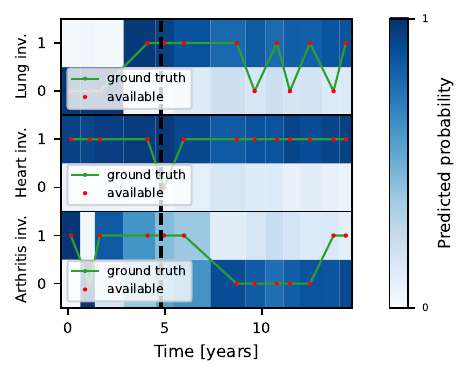}}\hfill%
        \subfigure[FVC: predicted mean and $95\%$ CIs at two different time points]{\label{fig:fvc}%
            \includegraphics[width=0.32\linewidth]{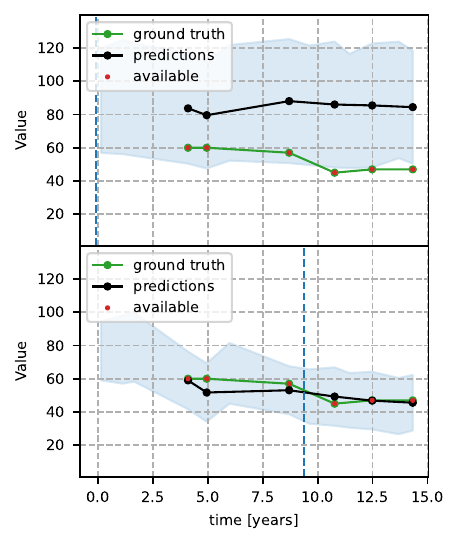}}
    }
\end{figure}

To illustrate how the model allows a holistic understanding of a patient's disease course, we follow an index patient $p_{\text{idx}}$ throughout the experiments. This patient has a complex disease trajectory, with varying organ involvement and stages.
We can use our model to forecast the high-dimensional distributions of $\bs{x}_{1:T}$ and $\bs{y}_{1:T}$ given the past measurements $\bs{x}_{0:k}$, as described in \autoref{sec:monit}.
The plots in \autoref{fig:monit_combined} show the predicted probabilities of organ involvement and predicted values of Forced Vital Capacity (FVC)\footnote{FVC is the amount of air that can be exhaled from the lungs.} at different time points for $p_{\text{idx}}$. The plots are overlaid with the ground truth labels in green. In particular, Figure \ref{fig:fvc} shows how the predictions become more accurate when more prior information is available to the model. 
We provide online prediction plots for additional $\bs{x}$ and $\bs{y}$ in Appendix \ref{sec:app:monit}.
% \begin{figure}[htbp]
% \centering
% \floatconts
% {fig:inv_pat}
% {\caption{Predicted probabilities of organ involvement for example patient $p_{\text{idx}}$, where values after the dashed line are forecasted.}}
% {\includegraphics[width=.9\linewidth]{graphs/inv_pat.pdf}}
% \end{figure}

% \begin{figure}[htbp]
% \centering
% \floatconts
% {fig:fvc}
% {\caption{FVC of $p_{\text{idx}}$: predicted mean and $95\%$ CI}}
% {\includegraphics[width=0.6\linewidth]{graphs/fvc_pat_2.pdf}}
% \end{figure}

% \begin{figure}[htbp]
%     \centering
%     \begin{subfigure}{\textwidth}
%         \centering
%         \floatconts{fig:inv_pat}{
%             \caption{Predicted probabilities of organ involvement for example patient $p_{\text{idx}}$, where values after the dashed line are forecasted.}
%         }{
%             \includegraphics[width=.9\linewidth]{graphs/inv_pat.pdf}
%         }
%     \end{subfigure}

%     \vspace{1em} % Add some vertical space between the subfigures

%     \begin{subfigure}{\textwidth}
%         \centering
%         \floatconts{fig:fvc}{
%             \caption{FVC of $p_{\text{idx}}$: predicted mean and $95\%$ CI}
%         }{
%             \includegraphics[width=0.6\linewidth]{graphs/fvc_pat_2.pdf}
%         }
%     \end{subfigure}
    
%     \caption{Combined figure with subfigures.}
% \end{figure}

In the next sections, we explore further applications and results of our model. While the performance was computed on validation sets, the subsequent results are derived from applying our model to a separate withheld test set. Furthermore, all of the $t-$SNE projections \citep{van2008visualizing} of the test set were obtained following the procedure described in Appendix \ref{sec:app:tsne}.

\subsection{Results: Cohort Analysis} 
By learning the joint distribution $p(\bs{x},\bs{y},\bs{z})$, our model allows us to analyze disease patterns in the cohort through the analysis of the latent process $\bs{z}$. Furthermore, by learning $p(\bs{z} \vert \bs{c})$, we estimate the average prior disease trajectories in the cohort. We analyze these prior trajectories in Appendix \ref{sec:app:prior}.
%By learning $p(\bs{x},\bs{y} \vert \bs{s}, \bs{\tau})$, we estimate the average prior disease trajectories in the cohort. 
% \begin{figure}[htbp]
%     \floatconts
%     {fig:guidance}
%     {\caption{Guided versus unguided latent spaces, overlaid with heart stage medical label.}}
%     {\includegraphics[width=0.55\linewidth]{graphs/guided_vs_unguided}}
% \end{figure}

% \begin{figure}[htbp]
%     \floatconts
%     {fig:tsnes}
%     {\caption{Probabilities of lung and heart involvement in the latent space.}}
%     {\includegraphics[width=0.55\linewidth]{graphs/lung_and_heart.png}}
% \end{figure}
\begin{figure}[htbp]
    \floatconts
    {fig:combined_figures}
    {\caption{Analysis of latent spaces.}}
    {%
        \subfigure[Guided versus unguided latent spaces, overlaid with heart stage medical label.]{%
            \label{fig:guidance}%
            \includegraphics[width=0.48\linewidth]{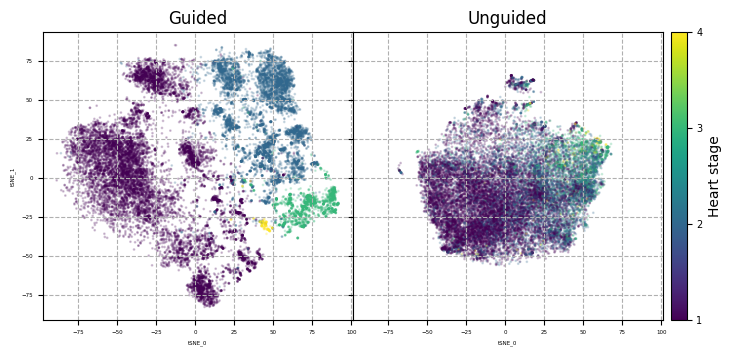}%
        }%
        \hfill%
        \subfigure[Probabilities of lung and heart involvement in the latent space.]{%
            \label{fig:tsnes}%
            \includegraphics[width=0.48\linewidth]{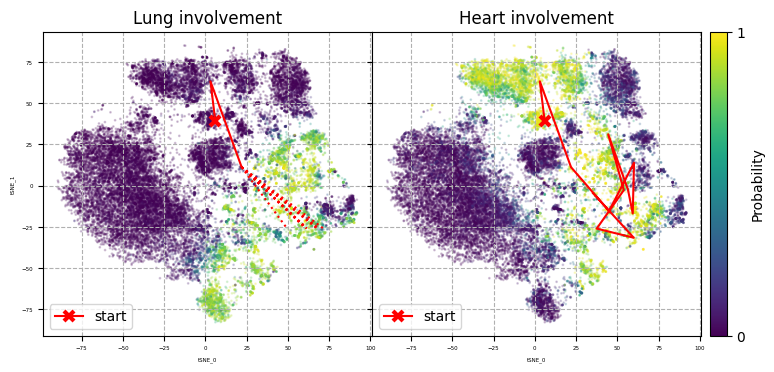}%
        }%
    }
\end{figure}

\subsubsection{Latent Space and Medical Labels}

% \begin{figure}[htbp]
%     \floatconts
%     {fig:tsnes}
%     {\caption{\textbf{Probabilities of lung and heart involvement} in the latent space.}}
%     {\includegraphics[width=\linewidth]{graphs/lung_and_heart.png}}
% \end{figure}
We aim to provide a method achieving semi-supervised disentanglement in the latent space. In \ref{fig:guidance}, we compare the distribution of the ground truth medical labels (here \emph{heart stage}) in a guided versus an unguided model (i.e.\ without training any guidance networks). The guided model clearly provides higher medical knowledge label disentanglement than the unguided model and thus enhances the interpretability of the different subspaces in $\bs{z}$. 
% Furthermore, this modeling approach allows for the analysis of organ-specific latent dimensions individually and thus aids in further understanding of organ-specific processes.

%trajectory in latent space 

In \ref{fig:tsnes}, we visualize the latent space overlaid with the different predicted probabilities of organ involvement. In red, we draw the latent space trajectory of $p_{\text{idx}}$, thus getting an understandable overview of their trajectory with respect to the different medical labels. The solid line highlights the reconstructed trajectory, whereas the dotted lines are forecasted sampled trajectories.  
% \begin{figure}[htbp]
%     \floatconts
%     {fig:tsnes}
%     {\caption{Probabilities of lung and heart involvement in the latent space.}}
%     {\includegraphics[width=0.55\linewidth]{graphs/lung_and_heart.png}}
% \end{figure}

In the first panel of \ref{fig:tsnes}, we leverage the model's generative abilities to sample forecasted $\bs{z}$ trajectories (dotted lines), providing estimates of future disease stages. The model forecasts that $p_{\text{idx}}$ will move towards a region with higher probabilities of lung and heart involvement. All of the sampled trajectories converge towards the same region in this case. The second panel is overlaid with the complete reconstructed trajectory of $p_{\text{idx}}$ in the latent space. The disentanglement in the latent space enables a straightforward overview of the past and future patient trajectory. 
Additionally, \autoref{fig:tsnes_stages} in the appendix shows the patient trajectory overlaid with the predicted organ stages.

\subsubsection{Clustering and Similarity of Patient Trajectories}

\label{sec:clustering}
As described in \autoref{subsec:patient_similrity}, we compute the
dynamic-time-warping similarity measure for the latent trajectories $\mathcal{H}_i
=
h(\mathcal{T}_i)$, and subsequently apply \emph{k-means} or \emph{k-nn} to respectively cluster the multivariate time series $\{ \mathcal{H}_i \}_{i=1}^N$ or find similar patient trajectories.
%  The partially observed and high-dimensional  disease trajectories
%  $\mathcal{T}_i$ can be projected to the corresponding
%  temporal hidden trajectories $\mathcal{H}_i
% =
% h(\mathcal{T}_i)$, as explained in 
% \autoref{subsec:patient_similrity}. For these latent trajectories, we compute the
% dynamical-time-warping (\emph{dtw}) similarity measure \citep{Muller2007DynamicWarping}, 
% %This can be used to define a patient similarity metric over trajectories
% % $
% % d_{\mathcal{T}}\left(
% % \mathcal{T}_i, \mathcal{T}_j
% % \right)
% % =
% % d_{\mathcal{H}}\left(
% % \mathcal{H}_i, \mathcal{H}_j
% % \right)
% % =
% % dtw(\mathcal{H}_i, \mathcal{H}_j)
% % $.
% %With this similarity measure, 
% %we 
% so that we can apply \emph{k-means} or \emph{k-nn} to respectively cluster the multivariate time series $\{ \mathcal{H}_i \}_{i=1}^N$ into a
% hierarchy of new subtypes or find similar patient trajectories. 
We used the library implemented by
\citet{tavenard2020tslearn}. 
 We focus here on the trajectory clustering results and refer the reader to Appendix \ref{sec:app:clust}
 for the patient similarity/nearest neighbor analysis. 
\begin{figure}[htbp]
    \floatconts
    {fig:clust_traj_all}
    {\caption{
    %Clustered trajectories in the latent space, overlaid with the predicted probabilities of organ involvement and severity stages.
    Clustering of latent trajectories and predicted probabilities of organ involvement for the mean cluster trajectories.}}
    {%
        \subfigure[Mean cluster trajectories in the latent space (starting at the cross \textbf{x}), overlaid with predicted probabilities of organ involvement and severity stages.]{%
            \label{fig:clust_traj}%
            \includegraphics[width=0.6\linewidth]{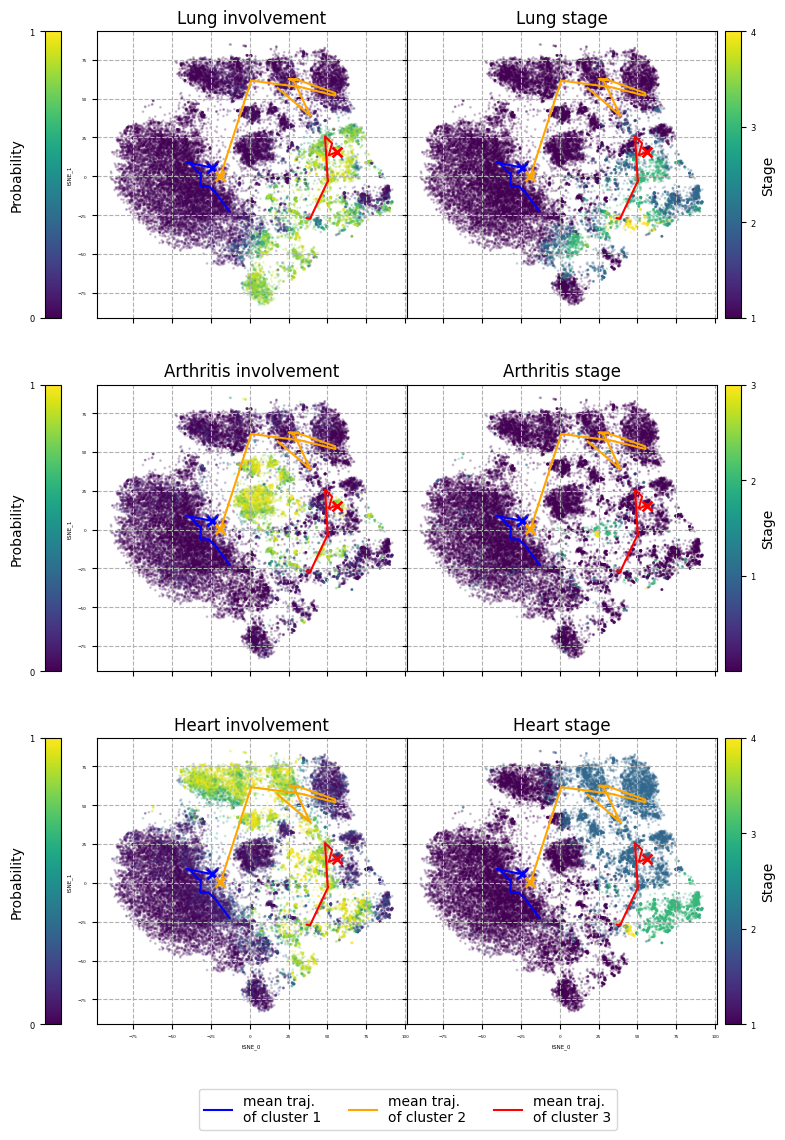}%
        }\hfill%
        \subfigure[Probabilities of organ involvement for cluster means.]{%
            \raisebox{0.8cm}{\label{fig:inv_probs}%
            \includegraphics[width=0.3\linewidth]{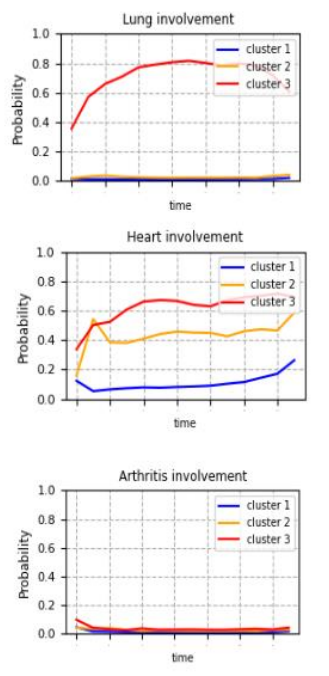}}%
        }%
    }
\end{figure}

% \begin{table*}[htbp]
%   \floatconts
%     {table:clust_prev} % label for cross-referencing
%  % table caption
%          {\caption{Prevalence of cutaneous involvement and gender in the clusters versus cohort prevalence. The arrows indicate the direction of the relative change compared to the cohort prevalence.}}
%     {
%     \begin{tabular}{lccc}  % adjust column alignment as needed
%       \toprule
%         & Diffuse Cutaneous Involvement of SSc & Male \\
%       \midrule
%       \textbf{Percentage in Cohort} &  \textbf{33\%} & \textbf{13\%} \\
%       Percentage in \textcolor{blue}{mild severity cluster} & 26\% $\downarrow$& 13\% \\
%       Percentage in \textcolor{orange}{medium severity cluster} &  31\% & 9\% $\downarrow$ \\
%       Percentage in \textcolor{red}{high severity cluster} & 46\% $\uparrow$ & 21\% $\uparrow$ \\
%       \bottomrule
%     \end{tabular}}

% \end{table*}

\begin{table*}[htbp]
  \floatconts
    {table:clust_prev} % label for cross-referencing
    {\caption{Prevalence of cutaneous involvement and gender in the clusters versus cohort prevalence. The arrows indicate the direction of the relative change compared to the cohort prevalence.}}
    {\small % This sets a smaller font size for the table
    \begin{tabular}{lccc}  % adjust column alignment as needed
      \toprule
        & Diffuse Cutaneous Involvement of SSc & Male \\
      \midrule
      \textbf{Percentage in Cohort} &  \textbf{33\%} & \textbf{13\%} \\
      Percentage in \textcolor{blue}{mild severity cluster} & 26\% $\downarrow$& 13\% \\
      Percentage in \textcolor{orange}{medium severity cluster} &  31\% & 9\% $\downarrow$ \\
      Percentage in \textcolor{red}{high severity cluster} & 46\% $\uparrow$ & 21\% $\uparrow$ \\
      \bottomrule
    \end{tabular}}
\end{table*}

\ref{fig:clust_traj} shows the three mean cluster trajectories in the latent space overlaid with the predicted medical labels. Moreover, we computed the predicted organ involvement probabilities for the cluster mean trajectories using the guidance networks (\ref{fig:inv_probs}). The first found cluster corresponds to patients with no or little organ involvement. The second mean trajectory starts close to the first but progresses towards regions with heart involvement. The third cluster contains the most severely progressing patients. 
The identified clusters show distinct patterns of organ involvement and disease severity (mild-medium-high), showing that the model separates disease trajectories into further subtypes.

\paragraph{Clustering Evaluation}
We evaluated our clustering approach both quantitatively and clinically. The optimal number of clusters $k$, was identified using the elbow method as shown in \autoref{fig:elbo} in the Appendix. We compared two methods: clustering latent trajectories $z$ against direct clustering of raw trajectories $x$. As discussed in Appendix \ref{sec:app:clust}, clustering latent trajectories achieves higher separation with respect to the medical labels compared to clustering the raw data. Lastly, contrarily to traditional clustering approaches, our method also supports \emph{predictive clustering}. Indeed, we can compare the cluster assignment of a forecasted trajectory (illustrated by the dotted line in \autoref{fig:combined_figures}) to the final cluster assignment based on the complete patient history encoded in the model. This approach achieved a macro $F_1$ score of $0.78$, indicating effective patient assignment to severity clusters early in their diagnosis.

We further evaluated the clinical relevance of the found clusters.
In \autoref{table:clust_prev}, we compare the prevalence of SSc subtypes (limited versus diffuse cutaneous SSc) and gender between the clusters. For instance, the cluster with the mildest severity contains a higher proportion of patients with limited cutaneous involvement at baseline, while the cluster with the highest severity includes a significantly larger number of patients with diffuse cutaneous SSc, showing that the model can separate the trajectories based upon the widely accepted SSc subtypes \citep{bains2017classification}. Furthermore, the most severe cluster exhibits a significantly higher proportion of male patients in comparison to the rest of the cohort. Recent studies also found that males tend to experience more frequent and severe lung and heart involvement, and concurrent organ involvement tends to result in poorer overall outcomes \citep{peoples2016gender, becker2019predictors}. 
 
\section{Discussion} 

% \emph{This is probably the most important section of your paper!  This
%   is where you tell us how your work advances our understanding of
%   machine learning and healthcare.}  Discuss both technical and
% clinical implications, as appropriate\cite{xyz19}.

In this paper, we present a novel deep semi-supervised generative latent variable approach to model complex disease trajectories. By introducing the guidance networks, we propose a method to augment unsupervised deep generative models with established medical knowledge and achieve more interpretable and disentangled latent processes. 

Our non-discriminative 
%generative approach 
approach
effectively addresses important desiderata for healthcare models such as forecasting, uncertainty quantification, dimensionality reduction, and interpretability. Furthermore, we empirically show that our model is suited for a real-world use case, namely the modeling of systemic sclerosis, and enables a holistic understanding of the patients' disease course.
%, and provides reliable and accurate predictions. 
The disentangled latent space facilitates comprehensive trajectory visualizations, straightforward analysis, and forecasting of patient trajectories. Most importantly, learning medically informed latent processes allows the discovery of novel clinically meaningful disease subtypes. We showed that the cluster separation is driven by clinically relevant features that have also been recognized as important predictors of SSc trajectories in recent studies.

\paragraph{Limitations and Future Work}

% Explain when your approach may not apply, or things you could not
% check.  \emph{Discussing limitations is essential.  Both ACs and
%   reviewers have been advised to be skeptical of any work that does
%   not consider limitations.}
While we have presented the benefits of proposing a multi-task ``holistic" model, this approach also has limitations. Naturally, the model is less performant at specific tasks compared to fine-tuned models, for instance, fully supervised predictive models for prediction. However, our modular approach could be adapted to excel in specific settings by removing certain components of the model.
%However, in this paper we aimed to propose a holistic approach, by emphasizing the unsupervised aspect and modeling jointly all of the involved covariates.

Our current approach holds the potential to be extended and adapted in several ways. We included only the most pertinent experiments and opted for a simple architecture suited to the modeling of systemic sclerosis. For instance, the model could be explicitly trained to reconstruct missing values, akin to denoising autoencoders. In future work, we intend to extend our framework to handle continuous time (Appendix \ref{sec:diff_prior}),
include medications for generating future hypothetical conditional trajectories (Appendix \ref{sec:app:genCondTraj}), 
include more organs in the modeling of SSc, and also include guidance networks to model additional disease dynamics like long-term outcomes. 
%In particular, the clinical insights regarding the knowledge discovery gained from the 
% clustering of trajectories such as finding new subtypes of SSc, identifying similar patients, and conducting a pathway analysis of the latent space, will be explored in future work modeling the involvement of additional organs. 

% ACKNOWLEDGEMENTS ONLY GO IN THE CAMERA-READY, NOT THE SUBMISSION
\acks{The authors thank the patients and caregivers who made the study possible, as well as all involved clinicians from the EUSTAR who collected the data. 
This work was funded by the Swiss National Science Foundation (project number 201184).}

\section*{Data and Code Availability}
The dataset used is owned by a third party, the EUSTAR group, and may be obtained by request after the approval and permission from EUSTAR.
The code builds upon the pythae library \citep{chadebec2022pythae}. 
The code and examples using some artificial data are available at \url{https://github.com/uzh-dqbm-cmi/eustar_mlhc}. 

%Do NOT change font size of references or modify the bibliography style
\bibliography{references}

\begin{thebibliography}{59}
\providecommand{\natexlab}[1]{#1}
\providecommand{\url}[1]{\texttt{#1}}
\expandafter\ifx\csname urlstyle\endcsname\relax
  \providecommand{\doi}[1]{doi: #1}\else
  \providecommand{\doi}{doi: \begingroup \urlstyle{rm}\Url}\fi

\bibitem[Alaa and van~der Schaar(2019)]{alaa2019attentive}
Ahmed~M Alaa and Mihaela van~der Schaar.
\newblock Attentive state-space modeling of disease progression.
\newblock \emph{Advances in neural information processing systems}, 32, 2019.

\bibitem[Allam et~al.(2021)Allam, Feuerriegel, Rebhan, and Krauthammer]{allam2021analyzing}
Ahmed Allam, Stefan Feuerriegel, Michael Rebhan, and Michael Krauthammer.
\newblock Analyzing patient trajectories with artificial intelligence.
\newblock \emph{Journal of medical internet research}, 23\penalty0 (12):\penalty0 e29812, 2021.

\bibitem[Bains(2017)]{bains2017classification}
Pooja Bains.
\newblock Classification criteria of systemic sclerosis: Journey so far.
\newblock \emph{Our Dermatology Online}, 8\penalty0 (2):\penalty0 220--223, 2017.

\bibitem[Baum and Petrie(1966)]{baum1966statistical}
Leonard~E Baum and Ted Petrie.
\newblock Statistical inference for probabilistic functions of finite state markov chains.
\newblock \emph{The annals of mathematical statistics}, 37\penalty0 (6):\penalty0 1554--1563, 1966.

\bibitem[Becker et~al.(2019)Becker, Graf, Sauter, Allanore, Curram, Denton, Khanna, Matucci-Cerinic, de~Oliveira~Pena, Pope, et~al.]{becker2019predictors}
Mike Becker, Nicole Graf, Rafael Sauter, Yannick Allanore, John Curram, Christopher~P Denton, Dinesh Khanna, Marco Matucci-Cerinic, Janethe de~Oliveira~Pena, Janet~E Pope, et~al.
\newblock Predictors of disease worsening defined by progression of organ damage in diffuse systemic sclerosis: a european scleroderma trials and research (eustar) analysis.
\newblock \emph{Annals of the rheumatic diseases}, 78\penalty0 (9):\penalty0 1242--1248, 2019.

\bibitem[Bengio et~al.(2013)Bengio, Courville, and Vincent]{bengio2013representation}
Yoshua Bengio, Aaron Courville, and Pascal Vincent.
\newblock Representation learning: A review and new perspectives.
\newblock \emph{IEEE transactions on pattern analysis and machine intelligence}, 35\penalty0 (8):\penalty0 1798--1828, 2013.

\bibitem[Bing et~al.(2021)Bing, Fortuin, and R{\"a}tsch]{bing2021disentanglement}
Simon Bing, Vincent Fortuin, and Gunnar R{\"a}tsch.
\newblock On disentanglement in gaussian process variational autoencoders.
\newblock \emph{arXiv preprint arXiv:2102.05507}, 2021.

\bibitem[Blei et~al.(2017)Blei, Kucukelbir, and McAuliffe]{blei2017variational}
David~M Blei, Alp Kucukelbir, and Jon~D McAuliffe.
\newblock Variational inference: A review for statisticians.
\newblock \emph{Journal of the American statistical Association}, 112\penalty0 (518):\penalty0 859--877, 2017.

\bibitem[Bonomi et~al.(2022)Bonomi, Peretti, Lepri, Venerito, Russo, Bruni, Iannone, Tangaro, Amedei, Guiducci, et~al.]{bonomi2022use}
Francesco Bonomi, Silvia Peretti, Gemma Lepri, Vincenzo Venerito, Edda Russo, Cosimo Bruni, Florenzo Iannone, Sabina Tangaro, Amedeo Amedei, Serena Guiducci, et~al.
\newblock The use and utility of machine learning in achieving precision medicine in systemic sclerosis: A narrative review.
\newblock \emph{Journal of Personalized Medicine}, 12\penalty0 (8):\penalty0 1198, 2022.

\bibitem[Casale et~al.(2018)Casale, Dalca, Saglietti, Listgarten, and Fusi]{casale2018gaussian}
Francesco~Paolo Casale, Adrian Dalca, Luca Saglietti, Jennifer Listgarten, and Nicolo Fusi.
\newblock Gaussian process prior variational autoencoders.
\newblock \emph{Advances in neural information processing systems}, 31, 2018.

\bibitem[Chadebec et~al.(2022)Chadebec, Vincent, and Allassonniere]{chadebec2022pythae}
Cl\'{e}ment Chadebec, Louis Vincent, and Stephanie Allassonniere.
\newblock Pythae: Unifying generative autoencoders in python - a benchmarking use case.
\newblock In S.~Koyejo, S.~Mohamed, A.~Agarwal, D.~Belgrave, K.~Cho, and A.~Oh, editors, \emph{Advances in Neural Information Processing Systems}, volume~35, pages 21575--21589. Curran Associates, Inc., 2022.

\bibitem[Chen et~al.(2021)Chen, Joshi, Ghassemi, and Ranganath]{chen2021probabilistic}
Irene~Y Chen, Shalmali Joshi, Marzyeh Ghassemi, and Rajesh Ranganath.
\newblock Probabilistic machine learning for healthcare.
\newblock \emph{Annual review of biomedical data science}, 4:\penalty0 393--415, 2021.

\bibitem[Chen et~al.(2018)Chen, Li, Grosse, and Duvenaud]{chen2018isolating}
Ricky~TQ Chen, Xuechen Li, Roger~B Grosse, and David~K Duvenaud.
\newblock Isolating sources of disentanglement in variational autoencoders.
\newblock \emph{Advances in neural information processing systems}, 31, 2018.

\bibitem[Chen et~al.(2023)Chen, Zheng, Mollaysa, Schürch, Allam, and Krauthammer]{chen2023dynamic}
Xingyu Chen, Xiaochen Zheng, Amina Mollaysa, Manuel Schürch, Ahmed Allam, and Michael Krauthammer.
\newblock Dynamic local attention with hierarchical patching for irregular clinical time series, 2023.

\bibitem[Cheng et~al.(2020)Cheng, Dumitrascu, Darnell, Chivers, Draugelis, Li, and Engelhardt]{cheng2020sparse}
Li-Fang Cheng, Bianca Dumitrascu, Gregory Darnell, Corey Chivers, Michael Draugelis, Kai Li, and Barbara~E Engelhardt.
\newblock Sparse multi-output gaussian processes for online medical time series prediction.
\newblock \emph{BMC medical informatics and decision making}, 20\penalty0 (1):\penalty0 1--23, 2020.

\bibitem[Chung et~al.(2015)Chung, Kastner, Dinh, Goel, Courville, and Bengio]{chung2015recurrent}
Junyoung Chung, Kyle Kastner, Laurent Dinh, Kratarth Goel, Aaron~C Courville, and Yoshua Bengio.
\newblock A recurrent latent variable model for sequential data.
\newblock \emph{Advances in neural information processing systems}, 28, 2015.

\bibitem[Comon(1994)]{comon1994independent}
Pierre Comon.
\newblock Independent component analysis, a new concept?
\newblock \emph{Signal processing}, 36\penalty0 (3):\penalty0 287--314, 1994.

\bibitem[Fortuin et~al.(2020)Fortuin, Baranchuk, R{\"a}tsch, and Mandt]{fortuin2020gp}
Vincent Fortuin, Dmitry Baranchuk, Gunnar R{\"a}tsch, and Stephan Mandt.
\newblock Gp-vae: Deep probabilistic time series imputation.
\newblock In \emph{International conference on artificial intelligence and statistics}, pages 1651--1661. PMLR, 2020.

\bibitem[Garaiman et~al.(2022)Garaiman, Nooralahzadeh, Mihai, Gonzalez, Gkikopoulos, Becker, Distler, Krauthammer, and Maurer]{garaiman2022vision}
Alexandru Garaiman, Farhad Nooralahzadeh, Carina Mihai, Nicolas~Perez Gonzalez, Nikitas Gkikopoulos, Mike~Oliver Becker, Oliver Distler, Michael Krauthammer, and Britta Maurer.
\newblock Vision transformer assisting rheumatologists in screening for capillaroscopy changes in systemic sclerosis: an artificial intelligence model.
\newblock \emph{Rheumatology}, page keac541, 2022.

\bibitem[Higgins et~al.(2016)Higgins, Matthey, Pal, Burgess, Glorot, Botvinick, Mohamed, and Lerchner]{higgins2016beta}
Irina Higgins, Loic Matthey, Arka Pal, Christopher Burgess, Xavier Glorot, Matthew Botvinick, Shakir Mohamed, and Alexander Lerchner.
\newblock beta-vae: Learning basic visual concepts with a constrained variational framework.
\newblock In \emph{International conference on learning representations}, 2016.

\bibitem[Hochreiter and Schmidhuber(1997)]{hochreiter1997long}
Sepp Hochreiter and J{\"u}rgen Schmidhuber.
\newblock Long short-term memory.
\newblock \emph{Neural computation}, 9\penalty0 (8):\penalty0 1735--1780, 1997.

\bibitem[Hoffmann-Vold et~al.(2021)Hoffmann-Vold, Allanore, Alves, Brunborg, Air{\'o}, Ananieva, Czirj{\'a}k, Guiducci, Hachulla, Li, et~al.]{hoffmann2021progressive}
Anna-Maria Hoffmann-Vold, Yannick Allanore, Margarida Alves, Cathrine Brunborg, Paolo Air{\'o}, Lidia~P Ananieva, L{\'a}szl{\'o} Czirj{\'a}k, Serena Guiducci, Eric Hachulla, Mengtao Li, et~al.
\newblock Progressive interstitial lung disease in patients with systemic sclerosis-associated interstitial lung disease in the eustar database.
\newblock \emph{Annals of the rheumatic diseases}, 80\penalty0 (2):\penalty0 219--227, 2021.

\bibitem[Holland et~al.(2023)Holland, Leingang, Holmes, Anders, Kaye, Riedl, Paetzold, Ezhov, Bogunovi{\'c}, Schmidt-Erfurth, et~al.]{holland2023clustering}
Robbie Holland, Oliver Leingang, Christopher Holmes, Philipp Anders, Rebecca Kaye, Sophie Riedl, Johannes~C Paetzold, Ivan Ezhov, Hrvoje Bogunovi{\'c}, Ursula Schmidt-Erfurth, et~al.
\newblock Clustering disease trajectories in contrastive feature space for biomarker discovery in age-related macular degeneration.
\newblock \emph{arXiv preprint arXiv:2301.04525}, 2023.

\bibitem[Hotelling(1933)]{hotelling1933analysis}
Harold Hotelling.
\newblock Analysis of a complex of statistical variables into principal components.
\newblock \emph{Journal of educational psychology}, 24\penalty0 (6):\penalty0 417, 1933.

\bibitem[Hsu et~al.(2017)Hsu, Zhang, and Glass]{hsu2017unsupervised}
Wei-Ning Hsu, Yu~Zhang, and James Glass.
\newblock Unsupervised learning of disentangled and interpretable representations from sequential data.
\newblock \emph{Advances in neural information processing systems}, 30, 2017.

\bibitem[Kim and Mnih(2018)]{kim2018disentangling}
Hyunjik Kim and Andriy Mnih.
\newblock Disentangling by factorising.
\newblock In \emph{International Conference on Machine Learning}, pages 2649--2658. PMLR, 2018.

\bibitem[Kingma and Ba(2014)]{kingma2014adam}
Diederik~P Kingma and Jimmy Ba.
\newblock Adam: A method for stochastic optimization.
\newblock \emph{arXiv preprint arXiv:1412.6980}, 2014.

\bibitem[Kingma and Welling(2013)]{kingma2013auto}
Diederik~P Kingma and Max Welling.
\newblock Auto-encoding variational bayes.
\newblock \emph{arXiv preprint arXiv:1312.6114}, 2013.

\bibitem[Lawley and Maxwell(1962)]{lawley1962factor}
David~N Lawley and Adam~E Maxwell.
\newblock Factor analysis as a statistical method.
\newblock \emph{Journal of the Royal Statistical Society. Series D (The Statistician)}, 12\penalty0 (3):\penalty0 209--229, 1962.

\bibitem[Lee and Van Der~Schaar(2020)]{lee2020temporal}
Changhee Lee and Mihaela Van Der~Schaar.
\newblock Temporal phenotyping using deep predictive clustering of disease progression.
\newblock In \emph{International conference on machine learning}, pages 5767--5777. PMLR, 2020.

\bibitem[Locatello et~al.(2020{\natexlab{a}})Locatello, Bauer, Lucic, R{\"a}tsch, Gelly, Sch{\"o}lkopf, and Bachem]{locatello2020sober}
Francesco Locatello, Stefan Bauer, Mario Lucic, Gunnar R{\"a}tsch, Sylvain Gelly, Bernhard Sch{\"o}lkopf, and Olivier Bachem.
\newblock A sober look at the unsupervised learning of disentangled representations and their evaluation.
\newblock \emph{The Journal of Machine Learning Research}, 21\penalty0 (1):\penalty0 8629--8690, 2020{\natexlab{a}}.

\bibitem[Locatello et~al.(2020{\natexlab{b}})Locatello, Poole, R{\"a}tsch, Sch{\"o}lkopf, Bachem, and Tschannen]{locatello2020weakly}
Francesco Locatello, Ben Poole, Gunnar R{\"a}tsch, Bernhard Sch{\"o}lkopf, Olivier Bachem, and Michael Tschannen.
\newblock Weakly-supervised disentanglement without compromises.
\newblock In \emph{International Conference on Machine Learning}, pages 6348--6359. PMLR, 2020{\natexlab{b}}.

\bibitem[Meier et~al.(2012)Meier, Frommer, Dinser, Walker, Czirjak, Denton, Allanore, Distler, Riemekasten, Valentini, et~al.]{meier2012update}
Florian~MP Meier, Klaus~W Frommer, Robert Dinser, Ulrich~A Walker, Laszlo Czirjak, Christopher~P Denton, Yannick Allanore, Oliver Distler, Gabriela Riemekasten, Gabriele Valentini, et~al.
\newblock Update on the profile of the eustar cohort: an analysis of the eular scleroderma trials and research group database.
\newblock \emph{Annals of the rheumatic diseases}, 71\penalty0 (8):\penalty0 1355--1360, 2012.

\bibitem[M{\"u}ller(2007)]{muller2007dynamic}
Meinard M{\"u}ller.
\newblock Dynamic time warping.
\newblock \emph{Information retrieval for music and motion}, pages 69--84, 2007.

\bibitem[Murphy(2022)]{murphy2022probabilistic}
Kevin~P Murphy.
\newblock \emph{Probabilistic machine learning: an introduction}.
\newblock MIT press, 2022.

\bibitem[Noroozizadeh et~al.(2023)Noroozizadeh, Weiss, and Chen]{noroozizadeh2023temporal}
Shahriar Noroozizadeh, Jeremy~C Weiss, and George~H Chen.
\newblock Temporal supervised contrastive learning for modeling patient risk progression.
\newblock In \emph{Machine Learning for Health (ML4H)}, pages 403--427. PMLR, 2023.

\bibitem[Palumbo et~al.(2023)Palumbo, Laguna, Chopard, and Vogt]{palumbo2023deep}
Emanuele Palumbo, Sonia Laguna, Daphn{\'e} Chopard, and Julia~E Vogt.
\newblock Deep generative clustering with multimodal variational autoencoders.
\newblock 2023.

\bibitem[Peoples et~al.(2016)Peoples, Medsger~Jr, Lucas, Rosario, and Feghali-Bostwick]{peoples2016gender}
Christine Peoples, Thomas~A Medsger~Jr, Mary Lucas, Bedda~L Rosario, and Carol~A Feghali-Bostwick.
\newblock Gender differences in systemic sclerosis: relationship to clinical features, serologic status and outcomes.
\newblock \emph{Journal of scleroderma and related disorders}, 1\penalty0 (2):\penalty0 204--212, 2016.

\bibitem[Poli{\v{c}}ar et~al.(2019)Poli{\v{c}}ar, Stra{\v{z}}ar, and Zupan]{polivcar2019opentsne}
Pavlin~G Poli{\v{c}}ar, Martin Stra{\v{z}}ar, and Bla{\v{z}} Zupan.
\newblock opentsne: a modular python library for t-sne dimensionality reduction and embedding.
\newblock \emph{BioRxiv}, page 731877, 2019.

\bibitem[Qin et~al.(2023)Qin, van~der Schaar, and Lee]{qin2023t}
Yuchao Qin, Mihaela van~der Schaar, and Changhee Lee.
\newblock T-phenotype: Discovering phenotypes of predictive temporal patterns in disease progression.
\newblock In \emph{International Conference on Artificial Intelligence and Statistics}, pages 3466--3492. PMLR, 2023.

\bibitem[Raghu et~al.(2023)Raghu, Chandak, Alam, Guttag, and Stultz]{raghu2023sequential}
Aniruddh Raghu, Payal Chandak, Ridwan Alam, John Guttag, and Collin Stultz.
\newblock Sequential multi-dimensional self-supervised learning for clinical time series.
\newblock In \emph{International Conference on Machine Learning}, pages 28531--28548. PMLR, 2023.

\bibitem[Ramchandran et~al.(2021)Ramchandran, Tikhonov, Kujanp{\"a}{\"a}, Koskinen, and L{\"a}hdesm{\"a}ki]{ramchandran2021longitudinal}
Siddharth Ramchandran, Gleb Tikhonov, Kalle Kujanp{\"a}{\"a}, Miika Koskinen, and Harri L{\"a}hdesm{\"a}ki.
\newblock Longitudinal variational autoencoder.
\newblock In \emph{International Conference on Artificial Intelligence and Statistics}, pages 3898--3906. PMLR, 2021.

\bibitem[Rosnati and Fortuin(2021)]{rosnati2021mgp}
Margherita Rosnati and Vincent Fortuin.
\newblock Mgp-atttcn: An interpretable machine learning model for the prediction of sepsis.
\newblock \emph{Plos one}, 16\penalty0 (5):\penalty0 e0251248, 2021.

\bibitem[Sch{\"u}rch et~al.(2020)Sch{\"u}rch, Azzimonti, Benavoli, and Zaffalon]{schurch2020recursive}
Manuel Sch{\"u}rch, Dario Azzimonti, Alessio Benavoli, and Marco Zaffalon.
\newblock Recursive estimation for sparse gaussian process regression.
\newblock \emph{Automatica}, 120:\penalty0 109127, 2020.

\bibitem[Sch{\"u}rch et~al.(2023{\natexlab{a}})Sch{\"u}rch, Azzimonti, Benavoli, and Zaffalon]{schurch2023correlated}
Manuel Sch{\"u}rch, Dario Azzimonti, Alessio Benavoli, and Marco Zaffalon.
\newblock Correlated product of experts for sparse gaussian process regression.
\newblock \emph{Machine Learning}, pages 1--22, 2023{\natexlab{a}}.

\bibitem[Sch{\"u}rch et~al.(2023{\natexlab{b}})Sch{\"u}rch, Li, Allam, Hofer, Mollaysa, Cavelti-Weder, and Krauthammer]{schurch2023generating}
Manuel Sch{\"u}rch, Xiang Li, Ahmed Allam, Giulia Hofer, Amina Mollaysa, Claudia Cavelti-Weder, and Michael Krauthammer.
\newblock Generating personalized insulin treatments strategies with conditional generative time series models.
\newblock In \emph{Deep Generative Models for Health Workshop NeurIPS 2023}, 2023{\natexlab{b}}.

\bibitem[Sch{\"u}rch(2022)]{schurch2022contributions}
Manuel~Pascal Sch{\"u}rch.
\newblock Contributions to scalable gaussian processes.
\newblock 2022.

\bibitem[Sohn et~al.(2015)Sohn, Lee, and Yan]{sohn2015learning}
Kihyuk Sohn, Honglak Lee, and Xinchen Yan.
\newblock Learning structured output representation using deep conditional generative models.
\newblock \emph{Advances in neural information processing systems}, 28, 2015.

\bibitem[S{\o}nderby et~al.(2016)S{\o}nderby, Raiko, Maal{\o}e, S{\o}nderby, and Winther]{sonderby2016ladder}
Casper~Kaae S{\o}nderby, Tapani Raiko, Lars Maal{\o}e, S{\o}ren~Kaae S{\o}nderby, and Ole Winther.
\newblock Ladder variational autoencoders.
\newblock \emph{Advances in neural information processing systems}, 29, 2016.

\bibitem[Srivastava and Rajan(2023)]{srivastava2023expertnet}
Shivin Srivastava and Vaibhav Rajan.
\newblock Expertnet: A deep learning approach to combined risk modeling and subtyping in intensive care units.
\newblock \emph{IEEE Journal of Biomedical and Health Informatics}, 2023.

\bibitem[Tavenard et~al.(2020)Tavenard, Faouzi, Vandewiele, Divo, Androz, Holtz, Payne, Yurchak, Ru{\ss}wurm, Kolar, et~al.]{tavenard2020tslearn}
Romain Tavenard, Johann Faouzi, Gilles Vandewiele, Felix Divo, Guillaume Androz, Chester Holtz, Marie Payne, Roman Yurchak, Marc Ru{\ss}wurm, Kushal Kolar, et~al.
\newblock Tslearn, a machine learning toolkit for time series data.
\newblock \emph{The Journal of Machine Learning Research}, 21\penalty0 (1):\penalty0 4686--4691, 2020.

\bibitem[Tomczak(2022)]{Tomczak2022DeepModeling}
Jakub~M. Tomczak.
\newblock {Deep Generative Modeling}.
\newblock \emph{Deep Generative Modeling}, pages 1--197, 1 2022.
\newblock \doi{10.1007/978-3-030-93158-2}.

\bibitem[Trottet et~al.(2023)Trottet, Allam, Micheroli, Horvath, Krauthammer, and Ospelt]{Trottet2023ExplainableDiseases}
Cécile Trottet, Ahmed Allam, Raphael Micheroli, Aron Horvath, Michael Krauthammer, and Caroline Ospelt.
\newblock {Explainable Deep Learning for Disease Activity Prediction in Chronic Inflammatory Joint Diseases}.
\newblock In \emph{ICML 3rd Workshop on Interpretable Machine Learning in Healthcare (IMLH)}, 2023.

\bibitem[Van Den~Oord et~al.(2017)Van Den~Oord, Vinyals, et~al.]{van2017neural}
Aaron Van Den~Oord, Oriol Vinyals, et~al.
\newblock Neural discrete representation learning.
\newblock \emph{Advances in neural information processing systems}, 30, 2017.

\bibitem[Van~der Maaten and Hinton(2008)]{van2008visualizing}
Laurens Van~der Maaten and Geoffrey Hinton.
\newblock Visualizing data using t-sne.
\newblock \emph{Journal of machine learning research}, 9\penalty0 (11), 2008.

\bibitem[Wang et~al.(2014)Wang, Sontag, and Wang]{wang2014unsupervised}
Xiang Wang, David Sontag, and Fei Wang.
\newblock Unsupervised learning of disease progression models.
\newblock In \emph{Proceedings of the 20th ACM SIGKDD international conference on Knowledge discovery and data mining}, pages 85--94, 2014.

\bibitem[Williams and Rasmussen(2006)]{williams2006gaussian}
Christopher~KI Williams and Carl~Edward Rasmussen.
\newblock \emph{Gaussian processes for machine learning}, volume~2.
\newblock MIT press Cambridge, MA, 2006.

\bibitem[Yang et~al.(2014)Yang, McAuley, Leskovec, LePendu, and Shah]{yang2014finding}
Jaewon Yang, Julian McAuley, Jure Leskovec, Paea LePendu, and Nigam Shah.
\newblock Finding progression stages in time-evolving event sequences.
\newblock In \emph{Proceedings of the 23rd international conference on World wide web}, pages 783--794, 2014.

\bibitem[Zhu et~al.(2022)Zhu, Xie, and Abd-Almageed]{zhu2022sw}
Jiageng Zhu, Hanchen Xie, and Wael Abd-Almageed.
\newblock Sw-vae: Weakly supervised learn disentangled representation via latent factor swapping.
\newblock In \emph{European Conference on Computer Vision}, pages 73--87. Springer, 2022.

\end{thebibliography}

\newpage
\appendix
% \section*{Appendix A.}

% Some more details about those methods, so we can actually reproduce
% them.  After the blind review period, you could link to a repository
% for the code also.  \emph{MLHC values both rigorous evaluation as well
%   as reproduciblity.}
\section{Systemic Sclerosis}
\subsection{Clinical Insights for Systemic Sclerosis}
 \label{se:clinic_ssc}

In this paper, we present a general approach for modeling and analyzing complex disease trajectories, for which we used
 the progression of systemic sclerosis as an example.
 The focus of this paper is on the machine learning methodology, while clinically relevant
insights and data analysis regarding systemic sclerosis
will be discussed in a clinical follow-up paper where
our model will be applied to investigate the involvement of multiple organs. 

Since there is ongoing research and discussion towards finding optimal definitions of the medical knowledge labels (involvement, stage, progression) for all impacted organs in SSc, we used preliminary definitions for three organs.

% \subsection{Dataset}
% \label{sec:app:data}

% The European Scleroderma Trials and Research group (EUSTAR) maintains a registry dataset of about 20’000 patients extensively documenting organ involvement in SSc. It contains around $30$ demographic variables, and 500 temporal clinical measurement variables documenting the patients' overall and organ-specific disease evolution. For a detailed description of the database, we refer the reader to \citet{meier2012update, hoffmann2021progressive}.

% For our analysis, we included $5673$ patients with enough temporality (i.e. at least $5$ medical visits). We used $10$ static variables related to the patients' demographics and $34$ clinical measurement variables, mainly related to the monitoring of the lung, heart, and joints in SSc.  In future work, we plan to include more patients and more clinical measurements for analyzing all involved organs.
\subsection{Medical Labels Definitions}
\label{sec:app:def}
Defining the organ involvement and stages in SSc is a challenging task as varying and sometimes contradicting definitions are used in different studies. However, there is ongoing research to find the most accurate definitions. Since this work is meant as a proof of concept, we used the following preliminary definitions of involvement and stage for the lung, heart, and joints (arthritis). The medical labels are defined for the variables of the EUSTAR database. There are $4$ stages of increasing severity for each organ. If multiple definitions are satisfied, the most severe stage is selected. Furthermore, there is missingness in the labels due to incomplete clinical measurements. Our modeling approach thus also could be used to label the medical labels when missing. 

We use the following abbreviations: 
\begin{itemize}
    \item Interstitial Lung Disease: ILD
    \item High-resolution computed tomography: HRCT
    \item Forced Vital Capacity: FVC
    \item Left Ventricular Ejection Fraction: LVEF
    \item Brain Natriuretic Peptide: BNP
    \item N-terminal pro b-type natriuretic peptide: NTproBNP
    \item Disease Activity Score 28: DAS28 
\end{itemize}
\subsubsection{Lung}
\paragraph{Involvement}
At least one of the following must be present: 
\begin{itemize}
    \item ILD on HRCT
    \item FVC $< 70 \%$
\end{itemize}
\paragraph{Severity staging}

\begin{enumerate}
    \item FVC $>80 \%$ or Dyspnea stage of $2$
    \item ILD extent $<20 \%$ or $70 \% < \text{FVC} \leq 80 \%$ or Dyspnea stage of $3$
    \item ILD extent $>20 \%$ or $50\% \leq \text{FVC} \leq 70 \%$ or Dyspnea stage of $4$
    \item FVC$<50\%$ or Lung transplant or Dyspnea stage of $4$
\end{enumerate}

\subsubsection{Heart}
\paragraph{Involvement}
At least one of the following must be present:
\begin{itemize}
    \item LVEF $<45\%$
    \item Worsening of cardiopulmonary manifestations within the last month
    \item Abnormal diastolic function
    \item Ventricular arrhythmias
    \item Pericardial effusion on echocardiography
    \item Conduction blocks
    \item BNP $>35$ pg/mL
    \item NTproBNP$>125$ pg/mL

\end{itemize}
\paragraph{Severity staging}
\begin{enumerate}
    \item Dyspnea stage of $1$
    \item Dyspnea stage of $2$
    \item Dyspnea stage of $3$
    \item Dyspnea stage of $4$
\end{enumerate}
\subsubsection{Arthritis}
\paragraph{Involvement}
At least one of the following must be present:
\begin{itemize}
    \item Joint synovitis
    \item Tendon friction rubs
\end{itemize}
\paragraph{Severity staging}
\begin{enumerate}
    \item DAS28 $<2.7$
    \item $2.7 \leq \text{DAS28} \leq 3.2$
    \item $3.2 < \text{DAS28} \leq 5.1$
    \item DAS28 $>5.1$
\end{enumerate}

\subsection{Model variables}
\label{app:model_variables}
Our model uses as temporal input features the following variables related to each organ and collected during medical visits: 
\begin{itemize}
\item Lung: 
Forced Vital Capacity, 
DLCO/SB, 
DLCOc/VA, 
Lung fibrosis,
Dyspnea (NYHA-stage),
Worsening of cardiopulmonary manifestations within the last month,
HRCT: Lung fibrosis,
Ground glass opacification,
Honey combing,
Tractions,
Reticular changes,
PAPsys (mmHg),
TAPSE: tricuspid annular plane systolic excursion in cm,
Right ventricular area (cm²) (right ventricular dilation),
Tricuspid regurgitation velocity (m/sec),
Pulmonary wedge pressure (mmHg),
Pulmonary resistance,
6 Minute walk test (distance in m)

\item Heart: Left ventricular ejection fraction,
Worsening of cardiopulmonary manifestations within the last month,
Diastolic function abnormal,
Ventricular arrhythmias,
Arrhythmias requiring therapy,
Pericardial effusion on echo,
Conduction blocks,
NTproBNP (pg/ml),
Auricular Arrhythmias,
BNP (pg/ml),
Cardiac arrhythmias,
Dyspnea (NYHA-stage)

\item Arthritis: Joint synovitis,
Joint polyarthritis,
Swollen joints,
Tendon friction rubs,
DAS 28 (ESR, calculated),
DAS 28 (CRP, calculated)

\end{itemize}

Moreover, we use the following the following (static) demographic variables: 
\begin{itemize}
\item Demographics: Sex, Height, Race, Subset of SSc according to LeRoy, Date of birth, Onset of first non-Raynaud's of the disease.
\end{itemize}

\section{Details and Extensions for Generative Model}
\label{subsec:gen_model_extension}
In this section, we provide  more details and several possible extensions to the main temporal generative model presented in Section
\ref{subsec:gen_model}.

\begin{table*}[ht]
\centering
\caption{Table of symbols}
\label{tab:symbols}
\begin{tabular}{cll}
\hline
\textbf{Symbol} & \textbf{Description} & \textbf{Domain} \\
\hline
$\bs{x} = \bs{x}_{1:T}$ & Clinical measurements (e.g., blood pressure). & $\mathbb{R}^{D \times T}$ \\
$\bs{y} = \bs{y}_{1:T}$ & Medical knowledge labels (e.g., disease severity). & $\mathbb{R}^{P \times T}$ \\
$\bs{\tau} = \bs{\tau}_{1:T}$ & Observation time-points. & $\mathbb{R}^{T}$ \\
$\bs{s}$ & Non-temporal info (e.g.\, patient demographics). & $\mathbb{R}^{S}$ \\
$\bs{p} = \bs{p}_{1:T}$ & Additional temporal covariates (e.g.\, medications). & $\mathbb{R}^{P \times T}$ \\
$\bs{c}$ & Context variables ($\bs{\tau}, \bs{p}, \bs{s}$). &  \\ 
$\bs{z}=\bs{z}_{1:T}$ & Multivariate latent processes & $\mathbb{R}^{L \times T}$ \\
$p_{\phi}(\bs{z} | \bs{c})$ & Prior network &  \\
$\phi$ & Prior network parameters  &  \\
$p_{\pi}(\bs{x} | \bs{z}, \bs{c})$ & Likelihood network &  \\
$\pi$ & Likelihood network parameters  &  \\
$p_{\gamma}(\bs{y} | \bs{z}, \bs{c})$ & Guidance network &  \\
$\gamma$ & Guidance network parameters &  \\
$\psi= \{\gamma, \pi, \phi\}$ & Parameters of the generative model. &  \\

$q_{\theta}(\bs{z} | \bs{x}_{0:k}, \bs{c})$ & Variational distribution  &  \\
$\theta$ & Variational parameters & \\
$\alpha, \beta$ & Weights in the training objective for balancing terms. & $\mathbb{R}_+$ \\
$\mathcal{T}$ & Observed patient trajectory (clinical measurements over time). & Sequence in $\mathbb{R}^{D \times T}$ \\
$\mathcal{H}$ & Latent patient trajectory (latent space representation). & Sequence in $\mathbb{R}^{L \times T}$ \\

\hline
\end{tabular}
\end{table*}

\subsection{Inference}
\label{sec:app:inference}
In this section, we explain the inference process of the proposed generative model $p_{\psi}(\bs{y}, \bs{x}, \bs{z} \vert \bs{c})=
p_{\gamma}(\bs{y} \vert \bs{z}, \bs{c})
       p_{\pi}(\bs{x} \vert \bs{z}, \bs{c})
      p_{\phi}(\bs{z} \vert \bs{c})$
in more detail.
We are particularly interested in the posterior of the latent variables $\bs{z}$
given  $\bs{y}$, $\bs{x}$, and $\bs{c}$, that is,
\begin{align*}
    p_{\psi}(\bs{z} \vert \bs{y}, \bs{x},  \bs{c})
    =
    \frac{p_{\psi}(\bs{y}, \bs{x}, \bs{z} \vert \bs{c})}
    {p_{\psi}(\bs{y}, \bs{x}\vert \bs{c})}
    =
    \frac{p_{\psi}(\bs{y}, \bs{x}, \bs{z} \vert \bs{c})}
    {\int p_{\psi}(\bs{y}, \bs{x}, \bs{z} \vert \bs{c}) d \bs{z} },
\end{align*}
which is in general intractable due to the marginalization of the latent process in the marginal likelihood 
$p_{\psi}(\bs{y}, \bs{x}\vert \bs{c}) = \int p_{\psi}(\bs{y}, \bs{x}, \bs{z} \vert \bs{c}) d \bs{z}$.
Therefore, we resort to approximate
inference, in particular, amortized variational inference (VI) \citep{blei2017variational}, where a variational
distribution $q_{\theta}(\bs{z} \vert \bs{x}, \bs{c})$ close to the true posterior distribution
$p_{\psi}(\bs{z} \vert \bs{x}, \bs{y}, \bs{c}) \approx q_{\theta}(\bs{z} \vert \bs{x}, \bs{c})$ is introduced. The similarity between these distributions is usually measured
in terms of KL divergence \citep{murphy2022probabilistic}, therefore, we aim to find parameters 
%$\theta^*$ of the distribution $q_{\theta^*}$
satisfying
$$
\theta^*, \psi^*
=
\argmin_{\theta, \psi} 
KL\left[
q_{\theta}(\bs{z} \vert \bs{x}, \bs{c})
~\vert\vert~
p_{\psi}(\bs{z} \vert \bs{x}, \bs{y}, \bs{c})
\right].
$$
This optimization problem is equivalent  \citep{murphy2022probabilistic} to maximizing a lower bound 
$\mathcal{L}( \psi, \theta; \bs{x},  \bs{y}, \bs{c} ) \leq
p_{\psi}(\bs{y}, \bs{x}\vert \bs{c})
$ 
to the intractable marginal likelihood, that is,
$$
\theta^*, \psi^*
=
\argmax_{\theta, \psi} 
\mathcal{L}( \psi, \theta; \bs{x},  \bs{y}, \bs{c} ).
$$
In particular, this lower bound equals
\begin{align*}
% p_{\psi}(\bs{y}, \bs{x}\vert \bs{c})
% &
% \geq
%     \mathcal{L}( \psi, \theta; \bs{x},  \bs{y}, \bs{c} )
%     \\
 \mathcal{L}
     =
    &\int
    q_{\theta}(\bs{z} \vert \bs{x}, \bs{c})
    \log
    \frac{p_{\psi}(\bs{y}, \bs{x}, \bs{z} \vert \bs{c})}
    {q_{\theta}(\bs{z} \vert \bs{x}, \bs{c})}
    d \bs{z}
    \\
    = 
    &
    \int
    q_{\theta}(\bs{z} \vert \bs{x}, \bs{c})
    \log
    \frac{p_{\gamma}(\bs{y} \vert \bs{z}, \bs{c})
       p_{\pi}(\bs{x} \vert \bs{z}, \bs{c})
      p_{\phi}(\bs{z} \vert \bs{c})}
    {q_{\theta}(\bs{z} \vert \bs{x}, \bs{c})}
    d \bs{z},
\end{align*}
which can be rearranged to
\begin{align*}
    \mathcal{L}
     =
     &~
     \mathbb{E}_{q_{\theta}(\bs{z} \vert \bs{x}, \bs{c})}\left[
 \log p_{\pi}(\bs{x} \vert \bs{z}, \bs{c}) 
 %p_{\gamma}(\bs{y} \vert \bs{z}, \bs{c})
  \right]
  \\
  +
  &~
 \mathbb{ E}_{q_{\theta}(\bs{z} \vert \bs{x}, \bs{c})}\left[
 \log 
 %p_{\pi}(\bs{x} \vert \bs{z}, \bs{c}) 
 p_{\gamma}(\bs{y} \vert \bs{z}, \bs{c})
  \right]
  \\
  -
  &~
 KL\left[
q_{\theta}(\bs{z} \vert \bs{x}, \bs{c} )
~\vert\vert~
p_{\phi}(\bs{z} \vert \bs{c})
  \right].
\end{align*}
For the Gaussian prior and approximate posterior described in
Section 
\ref{sec:priorL} and \ref{sec:post}, respectively, the KL-term can be computed analytically and efficiently
\citep{Tomczak2022DeepModeling}.
On the other hand, the expectations $ \mathbb{ E}_{q_{\theta}}$
can be approximated with a few Monte-Carlo samples
$\bs{z}^1,\ldots,\bs{z}^s,\ldots,\bs{z}^S
\sim
q_{\theta}(\bs{z} \vert \bs{x}, \bs{c})
$ leading to
\begin{align*}
 \mathbb{E}_{q_{\theta}(\bs{z} \vert \bs{x}, \bs{c})}\left[
 \log p_{\pi}(\bs{x} \vert \bs{z}, \bs{c}) 
 p_{\gamma}(\bs{y} \vert \bs{z}, \bs{c})
  \right]
  \\
  \approx
  \frac{1}{S}
  \sum_{s=1}^S
   \log p_{\pi}(\bs{x} \vert \bs{z}^s, \bs{c}) 
 p_{\gamma}(\bs{y} \vert \bs{z}^s, \bs{c}).
\end{align*}

\subsubsection{
%Inference with 
Partially Observed Data }
\label{sec:partially_obs}
The measurements 
$\bs{x}\in \RR{D \times T}$ and the labels  $\bs{y}\in \RR{P \times T}$ contain  many missing values. We define the  indices
$\bs{o}_x \in \RR{D \times T}$ and $\bs{o}_y \in \RR{P \times T}$ for which the observations are actually measured. Therefore, 
we compute the lower bound only on the observed variables, i.e.\
$\log p_{\psi}(\bs{x}^{\bs{o}_x}, \bs{y}^{\bs{o}_y} \vert \bs{c}) \geq \mathcal{L}( \psi, \theta; \bs{x}^{\bs{o}_x},  \bs{y}^{\bs{o}_y}, \bs{c} ) $,
as is similarly done by \citet{fortuin2020gp, ramchandran2021longitudinal}.
This then leads for instance to
\begin{align*} 
\mathbb{E}_{q_{\theta}(\bs{z} \vert \bs{x}, \bs{c})}\left[
 \log p_{\pi}(\bs{x}^{\bs{o}_x} \vert \bs{z}, \bs{c}) p_{\gamma}(\bs{y}^{\bs{o}_y} \vert \bs{z}, \bs{c})
 \right],
 \end{align*}
where the related
log-likelihood
$\log p_{\pi}(\bs{x}^{\bs{o}_x} \vert \bs{z}, \bs{c})
%  =
% \log \prod_{t=1}^T 
%    \prod_{D=1}^D
%  p_{\pi}(x_{t}^d \vert \bs{z}_{t}, \bs{c}_t)
 =
 \log \prod_{t,d \in \bs{o}_x}
 p_{\pi}(x_{t}^d \vert \bs{z}_{t}, \bs{c}_t)
 =
  \sum_{t,d \in \bs{o}_x}
 \log p_{\pi}(x_{t}^d \vert \bs{z}_{t}, \bs{c}_t)
 $
is only summed over the actually observed measurements. The same can be derived for the medical labels $\bs{y}^{\bs{o}_y} $.

\subsubsection{
Lower Bound for N Samples }
\label{sec:app:Nsamples}
Given a dataset with $N$ $\mathrm{iid}$ patients $\mathcal{D} = \{\mathcal{D}_i\}_{i=1}^N= \{\bs{x}_{1:T_i}^i, \bs{y}_{1:T_i}^i, \bs{c}_{1:T_i}^i\}_{i=1}^N$, 
the lower bound to the marginal log-likelihood is
$$
\log p_{\psi}(\mathcal{D}) 
=
\log \prod_{i=1}^N p_{\psi}(\mathcal{D}_i) 
\geq
\sum_{i=1}^N
\mathcal{L}( \psi, \theta; \bs{x}^i,  \bs{y}^i, \bs{c}^i ),
$$
which is maximized through stochastic optimization with mini-batches (\autoref{sec:inference}).
Moreover, 
suppose we have $T+1$ iid copies of the whole dataset $\{ \mathcal{D}^{k}\}_{k=0}^T$, then
\begin{align*}
\log p_{\psi}(\{ \mathcal{D}^{k}\}_{k=0}^T) 
=
\log \prod_{i=1}^N  \prod_{k=0}^T p_{\psi}(\mathcal{D}_i^k) 
\\
\geq
\sum_{i=1}^N
\sum_{k=0}^T
\mathcal{L}_k( \psi, \theta; \bs{x}^{i,k},  \bs{y}^{i,k}, \bs{c}^{i,k} ),
\end{align*}
where 
$\mathcal{L}_k( \psi, \theta; \bs{x}^{i,k},  \bs{y}^{i,k}, \bs{c}^{i,k} )$
is the lower bound obtained by plugging in the corresponding
approximate posterior
$q_{\theta}(\bs{z} \vert \bs{x}_{0:k}, \bs{c})$.

\subsubsection{Predictive Distributions}
\label{sec:app:predictive_distribution}

The predictive distributions for the measurement $\bs{x}_{1:T}$ and label trajectories $\bs{y}_{1:T}$ in
\autoref{sec:monit} can be obtained via a two-stage Monte-Carlo approach. For instance, we can sample from the distribution of the measurements 
%$\bs{x}_{1:T}$
\begin{align*}
&q_*
%_{\pi^*, \theta^*}
(\bs{x}_{1:T} \vert \bs{x}_{0:k}, \bs{c} ) 
\\
=
&\int
p_{\pi^*}(\bs{x}_{1:T} \vert \bs{z}_{1:T}, \bs{c} ) q_{\theta^*}(\bs{z}_{1:T} \vert \bs{x}_{0:k}, \bs{c} ) d \bs{z}
\end{align*}
by first sampling from the latent trajectories
$$
\bs{z}_{1:T}^1,\ldots,\bs{z}_{1:T}^s,\ldots\bs{z}_{1:T}^S
\sim
q_{\theta^*}(\bs{z}_{1:T} \vert \bs{x}_{0:k}, \bs{c} ) $$
given the current observed measurements
$ \bs{x}_{1:k}$.
In a second step, for each of the samples, we compute 
$$
\bs{x}_{1:T}^1,\ldots,\bs{x}_{1:T}^u,\ldots\bs{x}_{1:T}^U
\sim
p_{\pi^*}(\bs{x}_{1:T} \vert \bs{z}_{1:T}^s, \bs{c} )$$ to represent
the overall uncertainty of the measurement distribution. 

\subsection{Different Prior}
\label{sec:diff_prior}

The factorized prior described in \autoref{sec:priorL} can be extended to continuous time with Gaussian processes  (GPs) \citep{williams2006gaussian, schurch2020recursive, schurch2023correlated, schurch2022contributions}, as introduced 
 by
 \cite{casale2018gaussian, fortuin2020gp} in the unsupervised setting.
 In particular, 
 we can replace
 \begin{align*}
 p_{\phi}(\bs{z} \vert \bs{c})
&
=
 p_{\phi}(\bs{z}_{1:T} \vert \bs{c}_{1:T})
 =
 \prod_{t=1}^T 
  \prod_{l=1}^L
  p_{\phi}(\bs{z}_t^l \vert \bs{c}_t)
  \\
  &
  =
 \prod_{t=1}^T 
     \prod_{l=1}^L
   \mathcal{N}\left(
   \bs{z}_t^l \vert 
\mu_{\phi}^l(\bs{c}_t), \sigma^l_{\phi}(\bs{c}_t)
   \right),
 \end{align*}
 with
 \begin{align*}
% p_{\phi}(\bs{z} \vert \bs{c})
%&
%=
 p_{\phi}(\bs{z}_{1:T} \vert \bs{c}_{1:T})
 % =
 % \prod_{t=1}^T 
 %  \prod_{l=1}^L
 %  p_{\phi}(\bs{z}_t^l \vert \bs{c}_t)
  % \\
  %&
  =
     \prod_{l=1}^L
   \mathcal{GP}\left(
   \bs{z}^l \vert 
m_{\phi}^l(\bs{c}), k^l_{\phi}(\bs{c}, \bs{c}')
   \right)
 \end{align*}
 with a mean function $m_{\phi}^l(\bs{c})$ and kernel $k^l_{\phi}(\bs{c}, \bs{c}')$, to take into account all the probabilistic correlations occurring in continuous time. This leads to a \textit{stochastic} dynamic process, which theoretically matches the assumed disease process more adequately than a deterministic one. A further advantage is the incorporation of prior knowledge via the choice of the particular kernels for each latent process so that different characteristics such as long and small lengthscales, trends, or periodicity can be explicitly enforced in the latent space. 

\subsection{Conditional Generative Trajectory Generation}
\label{sec:app:genCondTraj}
Our generative approach is also promising for conditional generative trajectory sampling, in a similar spirit as proposed by \cite{schurch2023generating}. In particular, if we use medications as additional covariates 
$\bs{p}=\bs{p}_{1:T}=\{ \bs{p}_{0:k}, \bs{p}_{k+1:T}\}$ in our approximate posterior distribution
$q_{\theta}(\bs{z} \vert \bs{x}_{0:k}, \bs{c})=
q_{\theta}(\bs{z} \vert \bs{x}, \bs{\tau}, \bs{s}, \bs{p}_{0:k}, \bs{p}_{k+1:T})$
with $\bs{c} = \{\bs{\tau}, \bs{s}, \bs{p}\}$,
 the model can be used to sample future hypothetical trajectories 
 $\bs{x}_{k+1:T}$
 with 
 %the learned conditional generative distribution
 %and the likelihood with
%
 \begin{align*}
%&
q_*
%_{\pi^*, \theta^*}
(\bs{x}_{k+1:T}  \vert \bs{x}_{0:k}, \bs{\tau}, \bs{s}, \bs{p}_{0:k}, \bs{p}_{k+1:T} ) 
\\
=
%&
\int
p_{\pi^*}(\bs{x}_{k+1:T} \vert \bs{z}, \bs{\tau}, \bs{s}, \bs{p}_{0:k}, \bs{p}_{k+1:T} )
\\
%&
q_{\theta^*}(\bs{z} \vert \bs{x}_{0:k}, \bs{\tau}, \bs{s}, \bs{p}_{0:k}, \bs{p}_{k+1:T}) 
d \bs{z}
\end{align*}
 based
 on
 future query
 medications
  $\bs{p}_{k+1:T}$.

\section{Model Implementation}

\subsection{Model Architecture}
\label{sec:app:archi}
We describe the architecture and inputs/outputs of the different neural networks in our final model for SSc. For a patient with measurement time points $\bs{\tau}_{1:T}$ of the complete trajectory, the model input at time $t \in \bs{\tau}$ are the static variables $\bs{s}$, the clinical measurements $\bs{x}_{0:t}$, and the trajectory time points $\bs{\tau}$. Thus for SSc modeling, we have that $\bs{c}=\{ \bs{\tau}, \bs{s} \}$. The model $\mathcal{M}$ outputs the distribution parameters of the clinical measurements and the organ labels for all trajectory time points $\bs{\tau}$. Without loss of generality, we assume that $\bs{x}^{1:M}$ are continuous variables and $\bs{x}^{M+1:D}$ categorical, so that the model can be described as
\begin{align*}
        \mathcal{M}: \left(\bs{c},\bs{x}_{0:t}\right) \longrightarrow \\ \left(\hat{\bs{\mu}}_{1:T}^{x^{1:M}}(t), \hat{\bs{\sigma}}_{1:T}^{x^{1:M}}(t), \hat{\bs{\pi}}_{1:T}^{x^{M+1:D}}(t), \hat{\bs{\pi}}_{1:T}^{y}(t) \right).
\end{align*}
We explicitly include the dependencies to $t$ to emphasize that the parameters of the whole trajectory are estimated given the information up to time $t$.
\begin{itemize}
    \item \textbf{Prior network}: The prior is a multilayer perceptron (MLP). It takes as input $\bs{c}$ and outputs the estimated mean and variance of the prior latent distribution $\hat{\mu}_{1:T}^{prior}$ and $\hat{\sigma}_{1:T}^{prior}$.  
    \item \textbf{Encoder network} (posterior): The encoder contains LSTM layers followed by fully connected feed-forward layers. It takes as input $\bs{x}_{0:t}$ and $\bs{c}$ and outputs the estimated mean and standard deviation of the posterior distribution of the latent variables $\hat{\mu}_{1:T}^{post}(t)$ and $\hat{\sigma}_{1:T}^{post}(t)$, from which we sample the latent variables $\bs{z}_{1:T}(t)$ (complete temporal latent process) given the information up to $t$.  
    \item \textbf{Decoder network} (likelihood): The decoder is an MLP and takes as input the sampled latent variables $\bs{z}_{1:T}(t)$ and $\bs{c}$ and outputs the estimated means and standard deviations $\hat{\mu}_{1:T}^{x^{1:M}}(t)$ and $\hat{\sigma}_{1:T}^{x^{1:M}}(t)$ of the distribution of the continuous clinical measurements and class probabilities $\hat{\pi}_{1:T}^{x^{M+1:D}}(t)$ of the categorical measurements.
    \item \textbf{Guidance networks}: For each organ, we define one MLP guidance network per related medical label (involvement and stage). A guidance network for organ %$o$, 
    $o \in \mathcal{O}:= \{lung, heart, joints \}$ and related medical label 
    %$m$, 
    $m \in \{inv, stage \}$, takes as input the sampled latent variables $\bs{z}_{1:T}^{\epsilon(o(m))}(t)$ and outputs the predicted class probabilities  $\hat{\pi}_{1:T}^{y^{\nu(o(m))}}(t)$ of the labels, where $\nu(o(m))$ are the indices in $y$ related to the medical label $o(m)$, and $\epsilon(o(m))$ the indices in the latent space. 
\end{itemize} 

%Explicit encoder more? 

\subsection{Training Objective}

We follow the notation introduced in Section \ref{sec:method} and Appendix \ref{subsec:gen_model_extension}. To train the model to perform forecasting, for each patient, we augment the data by assuming $T+1$ \emph{iid} copies of the data $x$ and $y$ (see also \ref{sec:app:Nsamples}) and recursively try to predict the last $T - t$, $t=0,...,T$ clinical measurements and medical labels.
The total loss for a patient $p$ is
\begin{equation}
\mathcal{L}_p = \sum_{t=0}^{T} \mathcal{L}(t), 
\end{equation}
where 
\begin{align*}
       \mathcal{L}(t) := NLL \left(\hat{\mu} ^{x^{1:M}} (t), \hat{\sigma}^{x^{1:M}} (t), \bs{x}^{1:M} \right) \\
       + CE \left( \hat{\pi} ^ {x^{M+1:D}}(t), \bs{x}^{M+1:D} \right) \\ + \alpha * CE \left( \hat{\pi}^y(t), \bs{y} \right) \\ + \beta * KL \left( \hat{\mu}^{prior}, \hat{\sigma} ^{prior}, \hat{\mu}^{post}(t), \hat{\sigma}^{post}(t)\right),
\end{align*}
where $NLL$, $CE$ and $KL$ are the negative log-likelihood, cross-entropy and KL divergence, respectively. Further, $\alpha$ and $\beta$ are hyperparameters weighting the guidance and KL terms.

\subsubsection{Model Optimization}
\label{sec:app:optim}
We only computed the loss with respect to the available measurements. We randomly split the set of patients $\mathcal{P}$ into a train set $\mathcal{P}_{train}$ and test set $\mathcal{P}_{test}$  and performed $5-$fold CV with random search on $\mathcal{P}_{train}$ for hyperparameter tuning. Following the principle of empirical risk minimization, we trained our model to minimize the objective loss over $\mathcal{P}_{train}$, using the Adam \citep{kingma2014adam} optimizer with mini-batch processing and early stopping. 

\subsubsection{Architecture and Hyperparameters}
We tuned the dropout rate and the number and size of hidden layers using $5$-fold CV, and used a simple architecture for our final model. The posterior network contains a single lstm layer with hidden state of size $100$, followed by two fully connected layers of size $100$. The likelihood network contains two separate fully connected layers of size $100$, learning the mean and variances of the distributions separately. The guidance networks contain a single fully connected layer of size $40$ and the prior network a single fully connected layer of size $50$. We used batch normalization, ReLU activations, and a dropout rate of $0.1$. We set $\alpha = 0.2$ and $\beta = 0.01$. 
\section{Results}
\subsection{Model Evaluation}
\label{sec:app:res}
% We discuss the evaluation results for unguided models, medical label prediction, and uncertainty quantification. In \autoref{fig:perf_x_bsl}, we compare the performance of the clinical measurement $\bs{x}$ prediction of the different guided models versus their unguided counterparts (with the same number of latent processes). Note that these unguided models are optimal baselines for $\bs{x}$ prediction since they are not trained to predict $\bs{y}$ too. As \autoref{fig:perf_x_bsl} shows, the unguided models usually outperform the guided models, but the difference is not significant for the probabilistic models. Unsurprisingly, the best performing model is a deterministic unguided model, i.e. not trained to learn the $\bs{z}$ and $\bs{y}$ distributions.
% \begin{figure}[htbp]

% \floatconts
%   {fig:perf_x_bsl}
%   {\caption{Performance for $x$ prediction, guided versus unguided models.}}
%   {\includegraphics[width=\linewidth]{graphs/perf_x_bsl.pdf}}
% \end{figure}

% \autoref{fig:perf_y} shows the macro $F_1$ scores for the medical labels $\bs{y}$ prediction of the different models. The models with fixed likelihood variance generally slightly outperform the models with learned variance. All of the models outperform the individualized and cohort baselines. 
\subsubsection{Baselines}
\label{app:res:bsl}
As discussed in \autoref{seq:res:bsl}, we evaluated our model against temporal latent variable models optimized to predict either only $\bs{x}$ or $\bs{y}$ in a fully supervised way. We implemented probabilistic and deterministic variants of these models. Following the notations introduced in \autoref{subsec:gen_model} for the encoder networks (posterior), decoder and guidance networks, we can rewrite the objectives of the baselines as variants of the objective of our model objective described in Equation \eqref{eq:ELBO}. 

For the prediction of the clinical measurements, the LSTM-MLP-x* optimizes the objective  
\begin{equation*}
\begin{split}
 \mathbb{E}_{q_{\theta}(\bs{z} \vert \bs{x}_{0:k}, \bs{c})}\left[
 \log p_{\pi}(\bs{x} \vert \bs{z}, \bs{c}) 
  \right] \\
  -
 \beta ~KL\left[
q_{\theta}(\bs{z} \vert \bs{x}_{0:k}, \bs{c} )
~\vert\vert~
p_{\phi}(\bs{z} \vert \bs{c})
  \right]
  ,
  \end{split}
\end{equation*} 
and the LSTM-MLP-x 
\begin{equation*}
\begin{split}
 \mathbb{E}_{q_{\theta}(\bs{z} \vert \bs{x}_{0:k}, \bs{c})}\left[
 \log p_{\pi}(\bs{x} \vert \bs{z}, \bs{c}) 
  \right]
  \end{split}
\end{equation*} respectively. Similarly, for the prediction of the medical labels, the LSTM-MLP-y* optimizes
\begin{equation*}
        \begin{split}
 \mathbb{ E}_{q_{\theta}(\bs{z} \vert \bs{x}_{0:k}, \bs{c})}\left[
 \log 
 %p_{\pi}(\bs{x} \vert \bs{z}, \bs{c}) 
 p_{\gamma}(\bs{y} \vert \bs{z}, \bs{c})
  \right]
  \\
  -
 \beta ~KL\left[
q_{\theta}(\bs{z} \vert \bs{x}_{0:k}, \bs{c} )
~\vert\vert~
p_{\phi}(\bs{z} \vert \bs{c})
  \right]
  ,
   \end{split}
\end{equation*}

and the LSTM-MLP-y
\begin{equation*}
        \begin{split}
 \mathbb{ E}_{q_{\theta}(\bs{z} \vert \bs{x}_{0:k}, \bs{c})}\left[
 \log 
 %p_{\pi}(\bs{x} \vert \bs{z}, \bs{c}) 
 p_{\gamma}(\bs{y} \vert \bs{z}, \bs{c})
  \right].
   \end{split}
\end{equation*}
\subsubsection{Results}
\autoref{fig:perf_y} shows the prediction performance of the different models for each of the medical labels. The supervised models generally slightly outperform our model, and all temporal models greatly outperform the MLP and cohort baselines.

As discussed in the model evaluation, given the same model capacity, the LSTM-MLP baselines are expected to outperform our approach since they learn simpler tasks and fewer variables. \autoref{fig:perf_latent} shows the effect of reducing the dimension of the latent space of the deterministic LSTM-MLP-x baseline. Contrarily to our model, this baseline learns neither the variance of the latent variables nor the distribution of $\bs{y}$.  For continuous $\bs{x}$, the LSTM-MLP-x with five dimensions in the latent space performs similarly to our model. 
\begin{figure}[htbp]
\floatconts
  {fig:perf_y}
  {\caption{Performance for $\bs{y}$ prediction.}}
  {\includegraphics[width=0.5\linewidth]{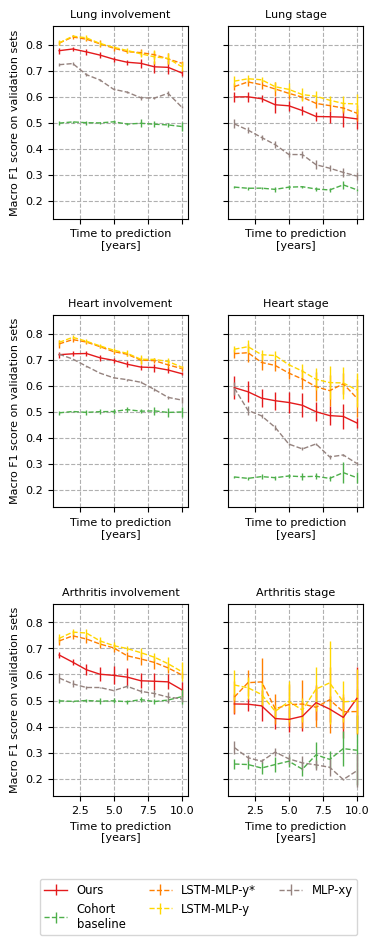}}
\end{figure}

\begin{figure}[htbp]
\floatconts
  {fig:perf_latent}
  {\caption{Effect of latent space dimension.}}
  {\includegraphics[width=0.5\linewidth]{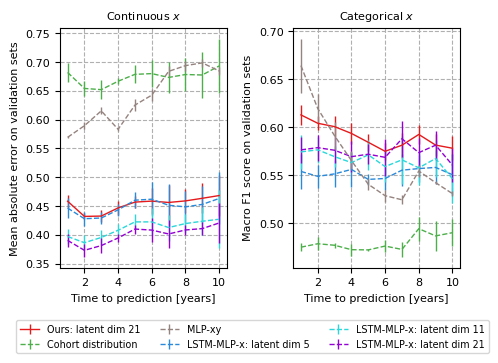}}
\end{figure}

\subsubsection{Uncertainty quantification}
To evaluate the uncertainty quantification of the models, we computed the coverage of the continuous predictions and calibration of the predicted probabilities for categorical measurements. The coverage is the probability that the confidence interval (CI) predicted by the model contains the true data point. Since the likelihood distribution is Gaussian, the 95\% CI is $\mu_{pred} \pm 1.96 \sigma_{pred}$. To achieve perfect coverage of the $95 \%$ CI, the predictions should fall within the predicted CI $95\%$ of the time. We computed the coverage over all forecasted data points. \autoref{fig:ood} shows the average ratio between CI length and feature range versus time to prediction. CIs are on average wider for long-term predictions and out-of-distribution data points, showing that the model predicts higher uncertainty for data points that are more difficult to predict.
For categorical measurements, the calibration curve is computed to assess the reliability of the predicted class probabilities. They are computed in the following way. We grouped all of the forecasted probabilities (for one-hot encoded vectors) into $n=20$ bins dividing the 0-1 interval. Then, for each bin, we compared the observed frequency of ground truth positives (aka ``fraction of positive") with the average predicted probability within the bin. Ideally, these two quantities should be as close as possible, i.e. close to the line of ``perfect calibration" in \autoref{fig:calib}.
The calibration curves in \autoref{fig:calib} show that all of the temporal models are well calibrated both in their categorical $\bs{x}$ and medical label $\bs{y}$ forecasts (averaged over all forecasted data points in the respective validation sets). 

\begin{figure}[htbp]
\floatconts
  {fig:calib}
  {\caption{Calibration curves for our model and the LSTM-MLPs.}}
  {\includegraphics[width=0.5\linewidth]{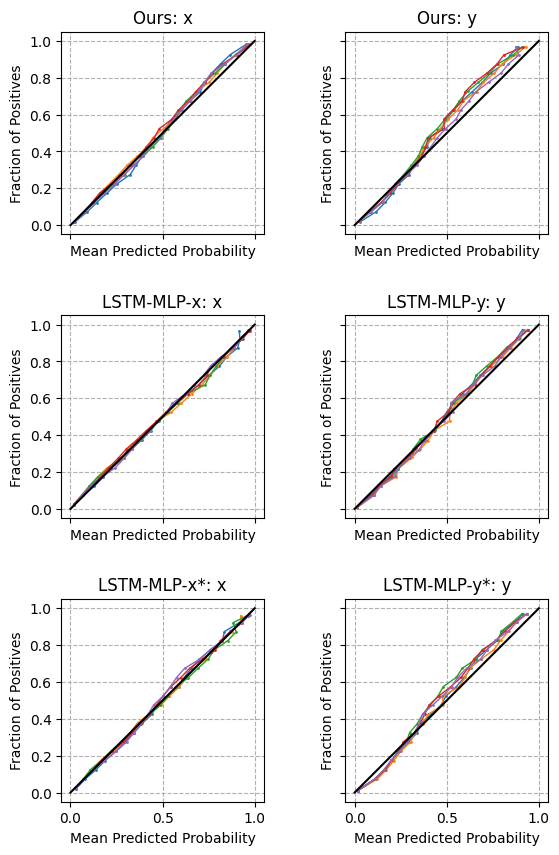}}
\end{figure}

\begin{figure}[htbp]
\floatconts
  {fig:ood}
  {\caption{Average ratio between CI length and feature range versus time to prediction. Out-of-distribution data points have wider CIs on average.}}
  {\includegraphics[width=0.35\linewidth]{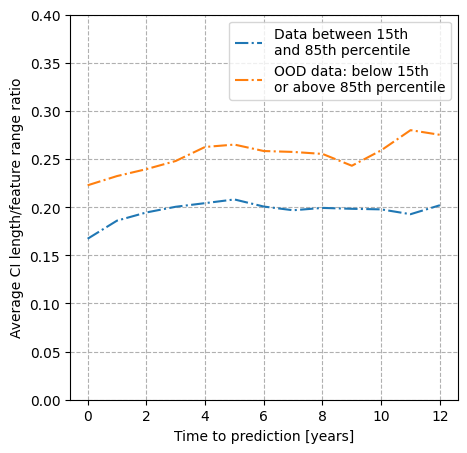}}
\end{figure}

\subsubsection{Online Prediction with Uncertainty}
\label{sec:app:monit}
We provide additional online prediction results for the index patient $p_{\text{idx}}$.

\begin{figure}[htbp]
\floatconts
  {fig:combined_pat}
  {\caption{DLCO(SB) and organ stage probabilities for $p_{\text{idx}}$.}}
  {%
    \subfigure[DLCO(SB) of $p_{\text{idx}}$: predicted mean and $95\%$ CI.]{\label{fig:dlco_pat}%
      \includegraphics[width=0.35\linewidth]{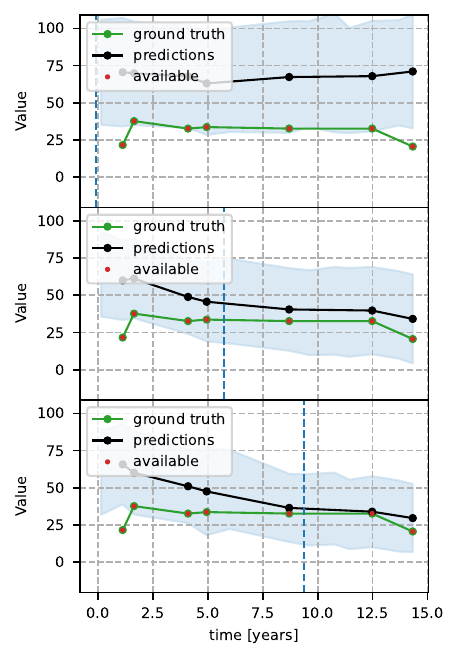}}%
    \hfill % Ensures that the subfigures are spaced out evenly across the available width
    \subfigure[Probabilities of organ stages for $p_{\text{idx}}$. The intensity of the heatmap reflects the predicted probability.]{\label{fig:stage_pat}%
      \includegraphics[width=0.35\linewidth]{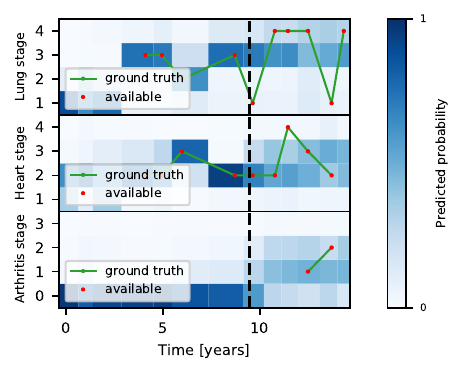}}
  }
\end{figure}

Figure \ref{fig:dlco_pat} shows the evolution in the predicted mean and $95\%$  CI of DLCO(SB)\footnote{DLCO(SB) stands for single breath (SB) diffusing capacity of carbon monoxide (DLCO).} for $p_{\text{idx}}$. 
The values after the dashed line are forecasted. As more prior information becomes available to the model, the forecast becomes more accurate and the CI shrinks. Moreover, in \autoref{fig:swollen_joints} we contrast the predicted uncertainty for a patient with an out-of-distribution (OOD) number of swollen joints (i.e.\ an unusually high number of swollen joints), and for the index patient. The model predicts significantly larger CIs for the OOD data point.
% \begin{figure}[htbp]
% \centering
% \floatconts
% {fig:fvc}
% {\caption{FVC of $p_{\text{idx}}$: predicted mean and $95\%$ CI}}
% {\includegraphics[width=0.6\linewidth]{graphs/fvc_pat_2.pdf}}
% \end{figure}
% \begin{figure}[htbp]

% \floatconts
%   {fig:dlco_pat}
%   {\caption{DLCO(SB) of $p_{\text{idx}}$: predicted mean and $95\%$ CI}}
%   {\includegraphics[width=0.5\linewidth]{graphs/dlcosb_pat.pdf}}
% \end{figure}
% Figure \autoref{fig:stage_pat} shows predicted probabilities of organ stages at a given time point. The intensity of the heatmap reflects the predicted probability.

% \begin{figure}[htbp]
% \floatconts{fig:stage_pat}
%   {\caption{Probabilities of organ stages for $p_{\text{idx}}$. }}
% {\includegraphics[width=0.5\linewidth]{graphs/stage_pat.pdf}}
% \end{figure}

\begin{figure}[htbp]
\floatconts
  {fig:swollen_joints}
  {\caption{Comparison between in and out-of-distribution predictions of swollen joints.}}
  {%
    \subfigure[Swollen joints for index patient]{\label{fig:swollen_pix}%
      \includegraphics[width=0.48\linewidth]{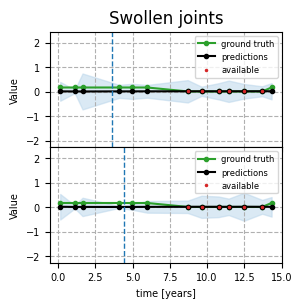}}%
    \hfill % Use \hfill to push the figures to the edge of the text area, making them side by side
    \subfigure[Swollen joints for out of distribution patient]{\label{fig:swollen_ood}%
      \includegraphics[width=0.46\linewidth]{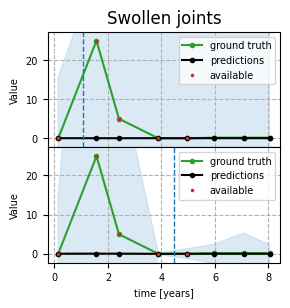}}
  }
\end{figure}
\subsection{Cohort Analysis}
We present here additional cohort-level experiments using our model. 
\subsubsection{Prior z Distributions}
% I will discuss this part a bit more, but not urgent
\label{sec:app:prior}

\begin{figure}[htbp]
\floatconts
  {fig:prior_pred}
  {\caption{Prior predicted $\bs{x}$ trajectories conditioned on time and static variables.}}
  {%
    \subfigure[Prior FVC trajectories overlaid with different SSC subsets.]{\label{fig:prior_fvc}%
      \includegraphics[width=0.49\linewidth]{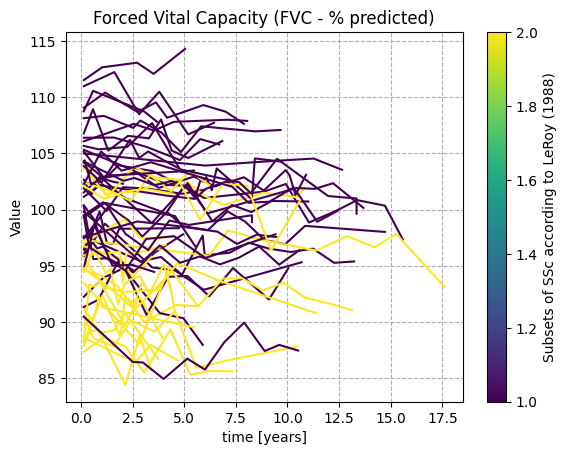}}%
    \hfill% This will fill the horizontal space between the two figures, making them appear side by side
    \subfigure[Prior natriuretic peptides trajectories overlaid with date of birth.]{\label{fig:ntb_prior}%
      \includegraphics[width=0.49\linewidth]{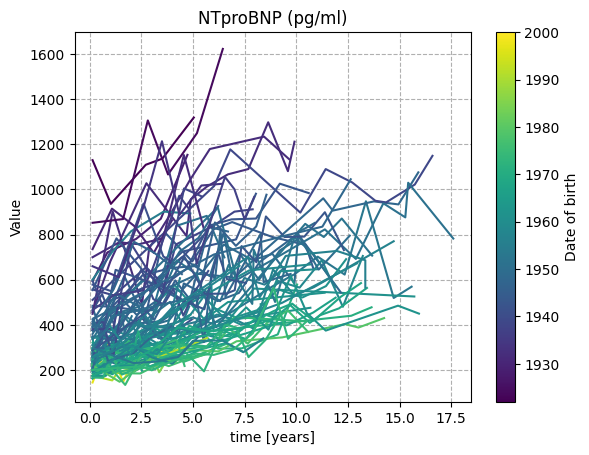}}
  }
\end{figure}

By learning $p(\bs{x},\bs{y} \vert \bs{s}, \bs{\tau})$, we estimate the average prior disease trajectories in the cohort. This allows the comparison of trajectories, conditioned only on the simple subset of variables $\bs{s}$ and $\bs{\tau}$ and thus without facing any confounding in the trajectories, for instance, due to past clinical measurements $\bs{x}$. For example, in  Figure \ref{fig:prior_fvc} we overlaid the predicted prior trajectories of Forced Vital Capacity (FVC)\footnote{FVC is the amount of air that can be exhaled from the lungs. Low levels indicate lung malfunction.} for a subset of patients in $\mathcal{P}_{test}$ with a static variable corresponding to the SSc subtype.  Overall, the FVC values are predicted to remain quite stable over time, but with different average values depending on the SSc subtype. In  Figure \ref{fig:ntb_prior}, the prior predicted N-terminal pro b-type natriuretic peptide (NTproBNP)\footnote{They are substances produced by the heart. High levels indicate potential heart failure.} trajectories overlaid with age, show that the model predicts an overall increase in NTproBNP over time, and steeper for older patients.
\subsubsection{Latent Space and Medical labels}
\begin{figure*}[htbp]
    \floatconts
    {fig:tsnes_stages}
    {\caption{Predicted organ stages in the latent space. The red line highlights the trajectory of $p_{\text{idx}}$.}}
    {\includegraphics[width=0.8\linewidth]{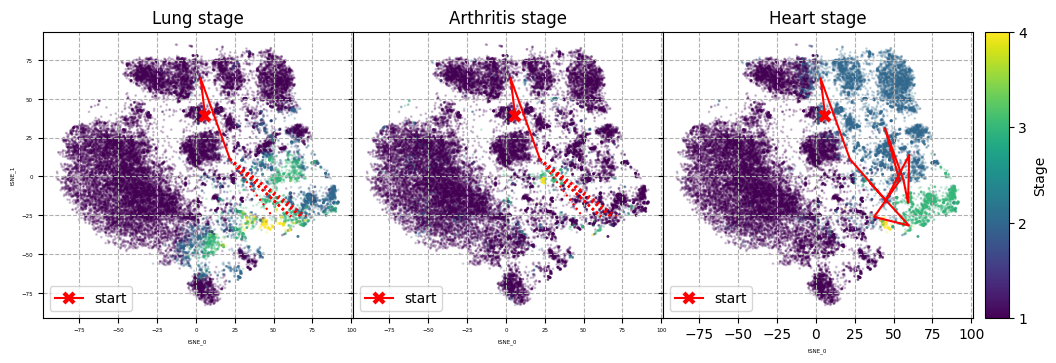}}
\end{figure*}
% \begin{table*}[htbp]
%   \floatconts
%     {table:clust_prev} % label for cross-referencing
%     {\caption{Prevalence of cutaneous involvement and gender in the clusters versus cohort prevalence. The arrows indicate the direction of the relative change compared to the cohort prevalence.}} % table caption
%     {
%     \begin{tabular}{lccc}  % adjust column alignment as needed
%       \toprule
%         & Diffuse Cutaneous Involvement of SSc & Male \\
%       \midrule
%       \textbf{Percentage in Cohort} &  \textbf{33\%} & \textbf{13\%} \\
%       Percentage in \textcolor{blue}{mild severity cluster} & 26\% $\downarrow$& 13\% \\
%       Percentage in \textcolor{orange}{medium severity cluster} &  31\% & 9\% $\downarrow$ \\
%       Percentage in \textcolor{red}{high severity cluster} & 46\% $\uparrow$ & 21\% $\uparrow$ \\
%       \bottomrule
%     \end{tabular}}
% \end{table*}
\label{sec:app:latent}
\paragraph{$\bf{t-}$SNEs:}{
\label{sec:app:tsne}
The $t$-SNE \citep{van2008visualizing} graphs were obtained by computing the two-dimensional $t$-SNE projection of the latent variables  $\bs{z}_{1:T} \mid (\bs{x}_{1:T}, \bs{c})$ (i.e. only using reconstructed $\bs{z}$) of a subset of $\mathcal{P}_{train}$ and then transforming and plotting the projected latent variables (reconstructed or forecasted) from patients in $\mathcal{P}_{test}$ \citep{polivcar2019opentsne}}.

In \ref{fig:tsnes}, we showed the trajectory of $p_{\text{idx}}$ overlaid with the predicted organ involvement probabilities. In \ref{fig:tsnes_stages}, we additionally show the trajectory overlaid with the organ stages, showing for instance in the first panel that the model predicts an increase in the lung stage and in the last panel that $p_{\text{idx}}$ undergoes many different heart stages throughout the disease course.

\subsubsection{Clustering of Patient Trajectories and Trajectory Similarity}
\label{sec:app:clust}

\begin{figure}[htbp]
\centering
\floatconts{fig:combined_med_label_traj}
  {\caption{Medical label trajectories.}}
  {%
    \subfigure[Medical label trajectories for cluster means.]{\label{fig:clust_traj_proba}%
      \includegraphics[width=0.48\linewidth]{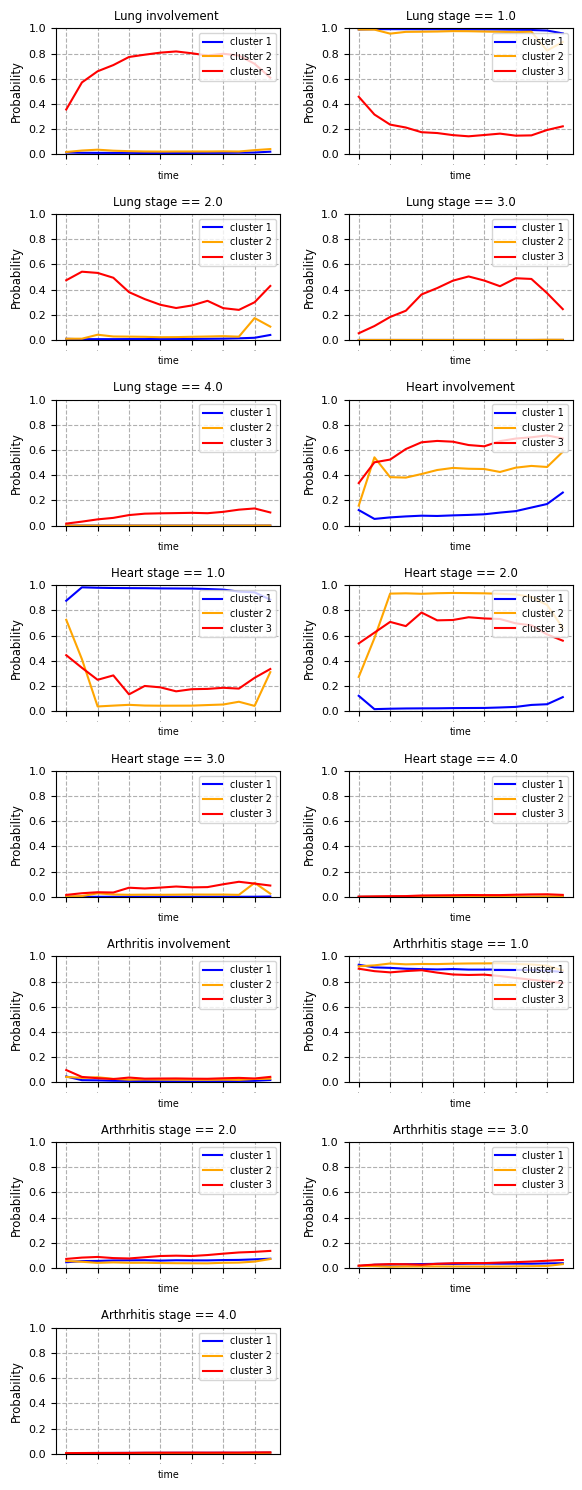}}%
    \hfill % This will space out the subfigures evenly
    \subfigure[Medical label trajectories for $p_{\text{idx}}$ and its $3$ nearest neighbors.]{\label{fig:traj_p_simil}%
      \includegraphics[width=0.48\linewidth]{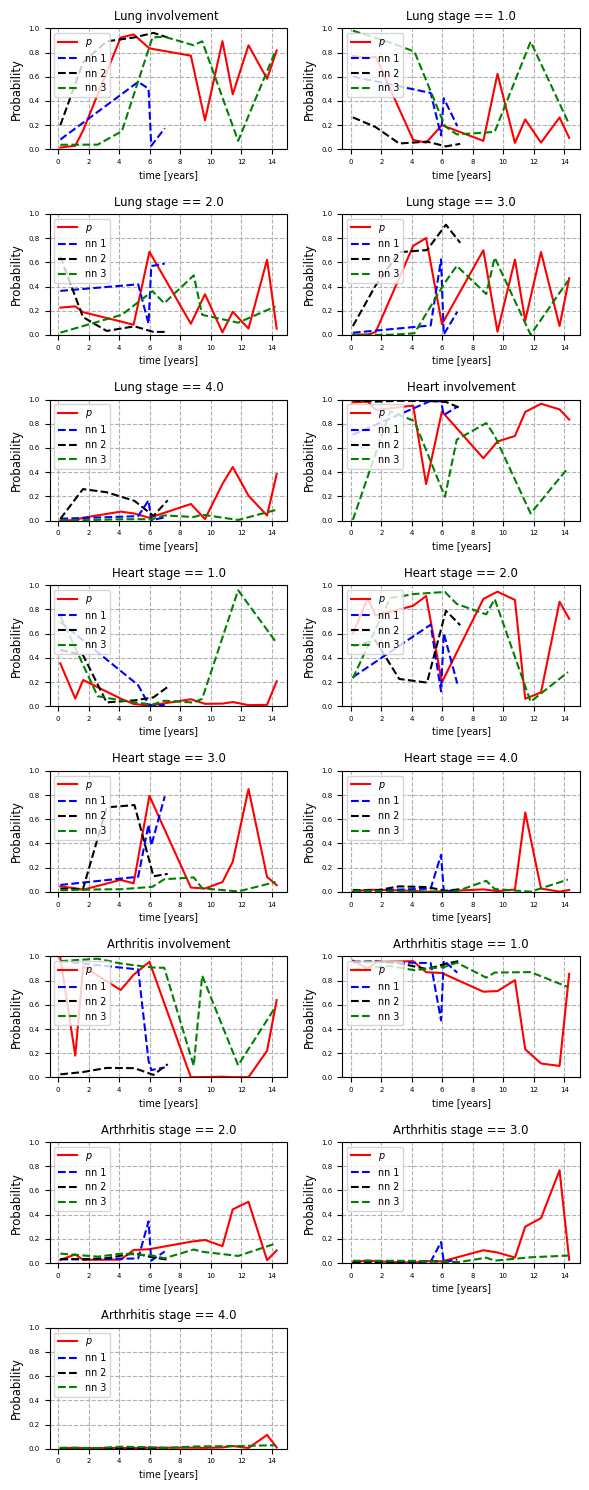}}
  }
\end{figure}

We discuss additional results obtained through clustering and similarity analysis of latent trajectories (\autoref{sec:clustering}). 
In \ref{fig:clust_traj_proba}, we show the different predicted probabilities of the medical labels $\bs{y}$ for the mean trajectories within the three found clusters. This reveals which medical labels are most differentiated by the clustering algorithm. For instance, cluster one exhibits low probabilities of organ involvement, while cluster two shows increasing probabilities of heart involvement and low probabilities of lung involvement. In contrast, cluster three shows increasing probabilities for both heart and lung involvement. We compared our approach of clustering latent trajectories $z$ to clustering the raw trajectories $x$ directly. In \autoref{fig:raw_vs_latent}, we compare the average medical label trajectories in the clusters obtained using both approaches. We see that clustering latent trajectories achieves more separation with respect to the medical labels than clustering the raw data. This indicates that our approach is better suited to uncover new subtypes with respect to medical knowledge. 
Furthermore, in \autoref{table:clust_prev}, we compare the prevalence of SSc subtypes (limited versus diffuse cutaneous SSc) and gender between the clusters. For instance, the most severe cluster contains an increased proportion of males compared to the cohort prevalence. 

\begin{figure*}[htbp]
\centering
\floatconts{fig:elbo}
  {\caption{Clustering: elbow plot for choice of optimal k. We set $k$ to $3$.}}
{\includegraphics[width=0.5\linewidth]{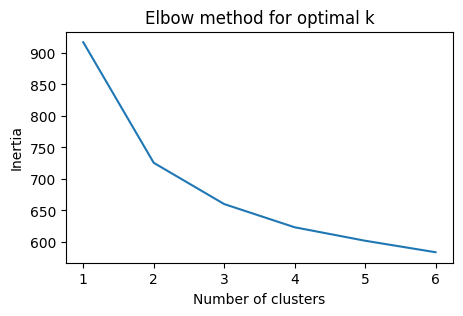}}
\end{figure*}

\begin{figure*}[htbp]
\centering
\floatconts{fig:raw_vs_latent}
  {\caption{Cluster separation with respect to medical labels. Comparison between clustering of raw $x$ trajectories versus clustering of latent trajectories $z$.  Our approach, where we cluster the latent trajectories, shows a higher separation with respect to the medical labels.}}
{\includegraphics[width=0.65\linewidth]{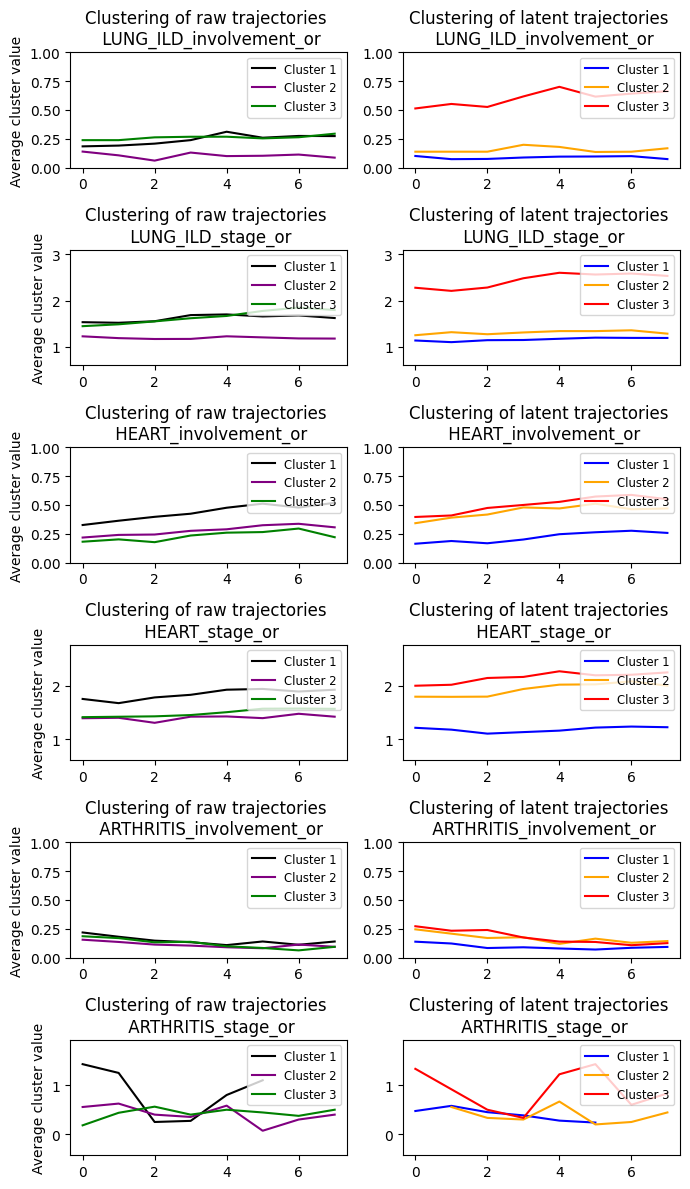}}
\end{figure*}

\begin{figure*}[htbp]
\centering
\floatconts{fig:tsne_traj_p_simil}
  {\caption{Trajectory of $p_{\text{idx}}$ and their $3$ nearest neighbors in the latent space. }}
{\includegraphics[width=0.8\linewidth]{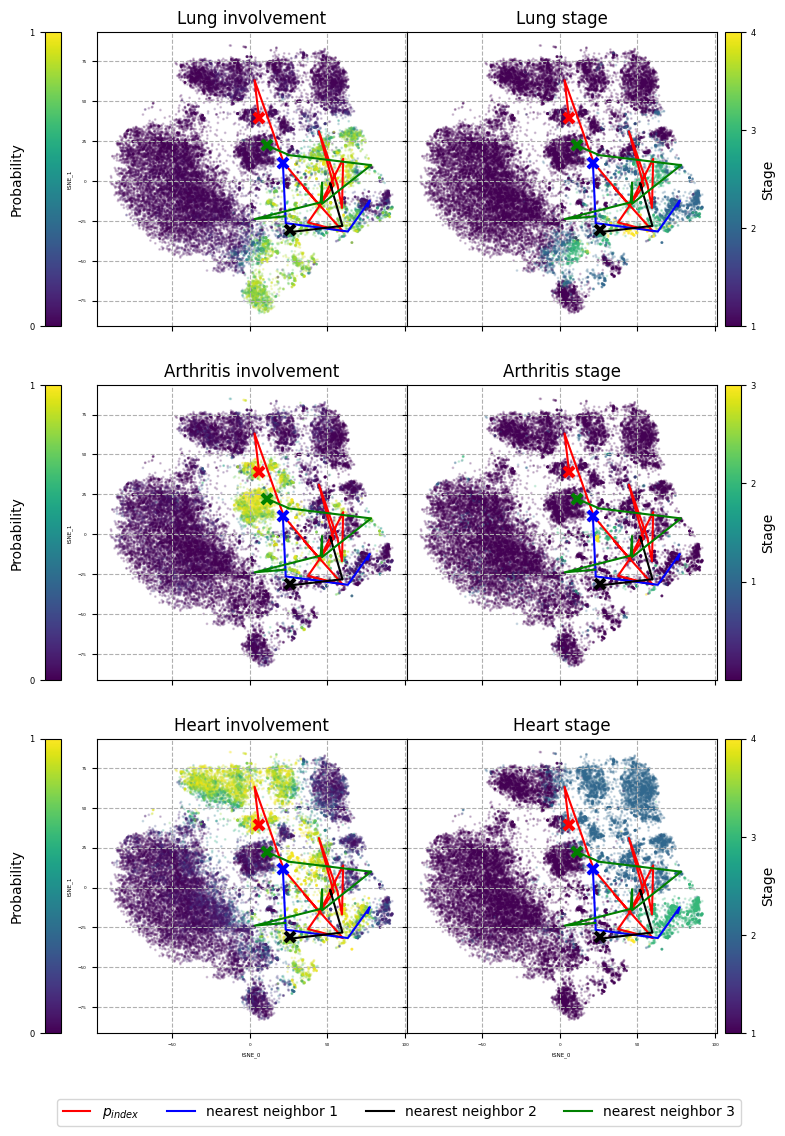}}
\end{figure*}
Additionally, we apply a \emph{k-nn} algorithm with the \emph{dtw} distance in the latent space to find patients with similar trajectories to $p_{\text{idx}}$. \autoref{fig:tsne_traj_p_simil} shows the trajectory of $p_{\text{idx}}$ and its three nearest neighbors in the latent space. We can see that the nearest neighbors also have an evolving disease, going through various organ involvements and stages. Similarly, in \ref{fig:traj_p_simil}, the medical label trajectories of $p_{\text{idx}}$ and its nearest neighbors reveal consistent patterns. 

% \begin{figure}[htbp]
% \centering
% \floatconts{fig:clust_traj_proba}
%   {\caption{Medical label trajectories for cluster means. }}
% {\includegraphics[width=0.5\linewidth]{graphs/traj_proba.png}}
% \end{figure}

% \begin{figure}[htbp]
% \centering
% \floatconts{fig:traj_p_simil}
%   {\caption{Medical label trajectories for $p_{\text{idx}}$ and its $3$ nearest neighbors. }}
% {\includegraphics[width=0.5\linewidth]{graphs/traj_p_simil.png}}
% \end{figure}

\end{document}